\documentclass[journal]{IEEEtran} 									

\newcommand{\Ncol}{two}   		

\usepackage[cmex10]{amsmath}
\usepackage{amssymb}
\usepackage{graphicx}
\usepackage{mathrsfs}
\usepackage{ifthen}
\usepackage{url,color,bbm}
\usepackage{multicol}
\usepackage[tight,footnotesize]{subfigure}

\graphicspath{{./Figures/}}

\def\L{{\rm L}}
\def\LI{{\rm L^{-1}}}

\def\D{{\rm D}}
\def\I{{\rm I}}
\def\SO{{\alpha}}
\def\ARC{{\lambda}}

\def\AI{{\theta}}
\def\d{{\rm d}}
\def\e{{\rm e}}
\def\j{{\rm j}}
\def\R{{\mathbb{R}}}
\def\FS{{\mathcal{E}}}
\def\CH{{\hat{\mathscr{P}}}}
\def\SP{{\beta_{{\rm L},T}}}

\def\EV{{\mathbb{E}}}
\def\bfOne{{\mathbbm{1}}}
\def\bbOne{{\mu}}
\def\bbTwo{{\sigma}}
\def\RP{{w}}
\def\IP{{IP}}

\def\ip{{ X}}
\def\ipl{{x}}

\newtheorem{defn}{\textbf{Definition}}
\newtheorem{lemma}{\textbf{Lemma}}
\newtheorem{theo}{\textbf{Theorem}}

\begin{document}

\title{Bayesian Estimation for Continuous-Time Sparse Stochastic Processes}

\author{Arash~Amini, 
			Ulugbek~S.~Kamilov,~\IEEEmembership{Student,~IEEE,}
        	Emrah~Bostan,~\IEEEmembership{Student,~IEEE,}
        	Michael~Unser,~\IEEEmembership{Fellow,~IEEE}
        	

\thanks{The authors are with the Biomedical Imaging Group (BIG), \'Ecole polytechnique f\'ed\'erale de Lausanne (EPFL), Lausanne, Switzerland. Emails: \{arash.amini,ulugbek.kamilov,emrah.bostan,michael.unser\}@epfl.ch.}

\thanks{This work was supported by the European Research Center (ERC) under FUN-SP grant.}
}



\maketitle

\begin{abstract}
We consider continuous-time sparse stochastic processes from which we have only a finite number of noisy/noiseless samples. Our goal is to estimate the noiseless samples (denoising) and  the signal in-between (interpolation problem).
 By relying on tools from the theory of splines, we derive the joint a priori distribution of the samples and show how this probability density function can be factorized. The factorization enables us to tractably implement the maximum a posteriori and minimum mean-square error (MMSE) criteria as two statistical approaches for estimating the unknowns. We compare the derived statistical methods with well-known techniques for the recovery of sparse signals, such as the $\ell_1$ norm and Log ($\ell_1$-$\ell_0$ relaxation) regularization methods. The simulation results show that, under certain conditions, the performance of the regularization techniques can be very close to that of the MMSE estimator.
\end{abstract}

\begin{IEEEkeywords}
Denoising, Interpolation, L\'evy Process, MAP, MMSE, Statistical Learning, Sparse Process.
\end{IEEEkeywords}

%
\IEEEpeerreviewmaketitle


\section{Introduction}

\IEEEPARstart{T}{he} recent popularity of regularization techniques in signal and image processing is motivated by the sparse nature of real-world data. It has resulted in the development of powerful tools for many problems such as denoising, deconvolution, and interpolation. The emergence of compressed sensing, which focuses on the recovery of sparse vectors from highly under-sampled sets of measurements, is playing a key role in this context \cite{Candes2006a,Candes2006b,Donoho2006}.

Assume that the signal of interest $\{s[i]\}_{i=0}^m$ is a finite-length discrete signal also represented by $\mathbf{s}$ as a vector) that has a sparse or almost sparse representation in some transform or analysis domain (e.g., wavelet or DCT). Assume moreover that we only have access to noisy measurements of the form $\big\{\tilde{s}[i]=s[i]+n[i]\big\}_{i=0}^m$, where $\big\{n[i]\big\}_{i=0}^m$ denotes an additive white Gaussian noise. Then, we would like to estimate $\{s[i]\}_i$. The common sparsity-promoting variational techniques rely on penalizing the sparsity in the transform/analysis domain \cite{Starck2005,Wright2009} by imposing
\begin{eqnarray}\label{eq:PenaltyDenoising}
\big\{\hat{s}[i]\big\}_{i=0}^m=\textrm{arg}\min_{\{s[i]\}}\big\{\|\mathbf{s}-\tilde{\mathbf{s}}\|_{\ell_2}^2+\lambda J_{\rm sparse}(\mathbf{s})\big\},
\end{eqnarray}
where $\mathbf{\tilde{s}}$ is the vector of noisy measurements, $J_{\rm sparse}(\cdot)$ is a penalty function that reflects the sparsity constraint in the transform/analysis domain and $\lambda$ is a weight that is usually set based on the noise and signal powers. The choice of $J_{\rm sparse}(\cdot)=\|\cdot\|_{\ell_1}$ is one of the favorite ones in compressed sensing when $\{s[i]\}_{i=0}^m$ is itself sparse \cite{Candes2005}, while the use of $J_{\rm sparse}(\mathbf{s})=TV(\mathbf{s})$, where $TV$ stands for total variation, is a common choice for piecewise-smooth signals that have sparse derivatives \cite{Rudin92}.

Although the estimation problem for a given set of measurements is a deterministic procedure and can be handled without recourse to statistical tools, there are benefits in viewing the problem from the stochastic perspective. For instance, one can take advantage of side information about the unobserved data to establish probability laws for all or part of the data. Moreover, a stochastic framework allows one to evaluate the performance of estimation techniques and argue about their distance from the optimal estimator.
The conventional stochastic interpretation of the variational method in (\ref{eq:PenaltyDenoising}) leads to the finding that $\{\hat{s}[i]\}_{i=0}^m$ is the maximum a posteriori (MAP) estimate of $\{s[i]\}_{i=0}^m$. In this interpretation, the quadratic data term is associated with the Gaussian nature of the additive noise, while the sparsifying penalty term corresponds to the a priori distribution of the sparse input. For example, the penalty $J_{\rm sparse}(\cdot)=\|\cdot\|_{\ell_1}$ is associated with the MAP estimator with Laplace prior \cite{Wipf2004,Park2008}. However, investigations of the compressible/sparse priors have revealed that the Laplace distribution cannot be considered as a sparse prior \cite{Cevher2008conf,Amini2011}, \cite{Gribonval2012}. Recently in \cite{Gribonval2011a}, it is argued that (\ref{eq:PenaltyDenoising}) is better interpreted as the minimum mean-square error (MMSE) estimator of a sparse prior.

Though the discrete stochastic models are widely adopted for sparse signals, they only approximate the continuous nature of real-world signals. The main challenge for employing continuous models is to transpose the compressibility/sparsity concepts in the continuous domain while maintaining compatibility with the discrete domain. In \cite{Unser2011}, an extended class of piecewise-smooth signals is proposed as a candidate for continuous stochastic sparse models. This class is closely related to signals with a finite rate of innovation \cite{Vetterli2022}. Based on infinitely divisible distributions, a more general stochastic framework has been recently introduced in \cite{Unser2011_P1,Unser2011_P2}. There, the continuous models include Gaussian processes (such as Brownian motion), piecewise-polynomial signals, and $\alpha$-stable processes as special cases. In addition, a large portion of the introduced family is considered as compressible/sparse with respect to the definition in \cite{Amini2011} which is compatible with the discrete definition.

In this paper, we investigate the estimation problem for the samples of the continuous-time sparse models introduced in \cite{Unser2011_P1,Unser2011_P2}. We derive the a priori and a posteriori probability density functions (pdf) of the noiseless/noisy samples.
 We present a practical factorization of the prior distribution which enables us to perform statistical learning for denoising or interpolation problems. In particular, we implement the optimal MMSE estimator based on the message-passing algorithm.  The implementation involves discretization and convolution of pdfs, and is in general, slower than the common variational techniques. We further compare the performance of the Bayesian and variational denoising methods. Among the variational methods, we consider quadratic, TV, and Log regularization techniques. Our results show that, by matching the regularizer to the statistics, one can almost replicate the MMSE performance.

The rest of the paper is organized as follows: In Section \ref{sec:SignalModel}, we introduce our signal model which relies on the general formulation of sparse stochastic processes proposed in \cite{Unser2011_P1,Unser2011_P2}. In Section \ref{sec:Estimation}, we explain the techniques for obtaining the probability density functions and, we derive the estimation methods in Section \ref{subsec:aPriorDist}. We study the special case of L\'evy processes which is of interest in many applications in Section \ref{sec:Levy}, and present simulation results in Section \ref{sec:SimulationResults}. Section \ref{sec:Conc} concludes the paper.


\begin{figure*}[tb]
\centering
\includegraphics[width=12.5cm]{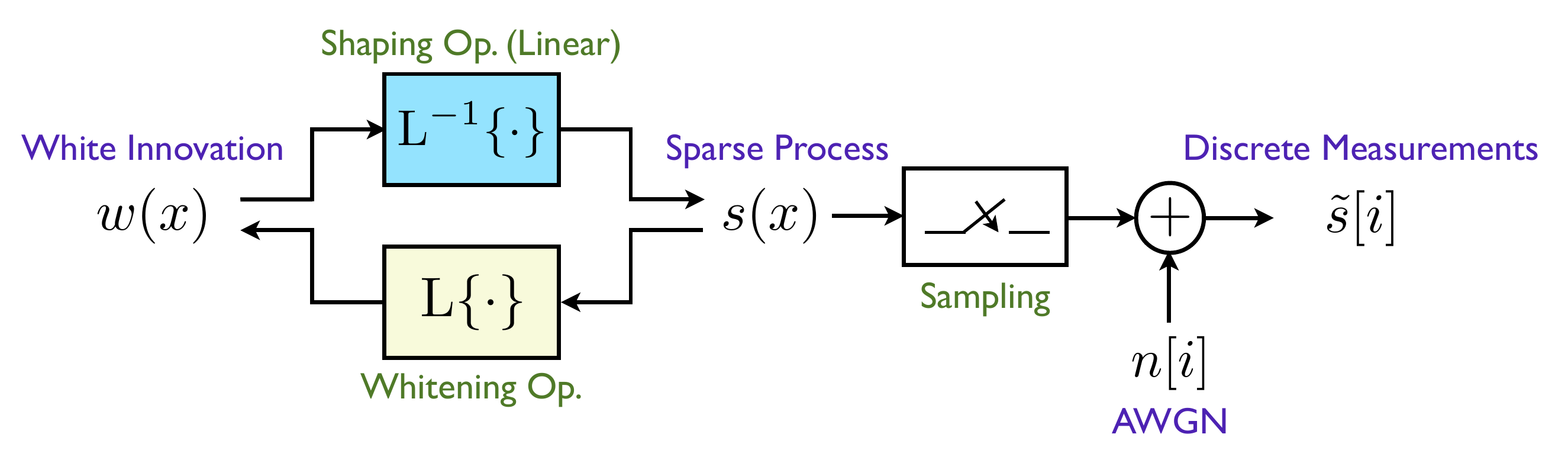}
\caption{Connection between the white noise $w(x)$, the sparse continuous signal $s(x)$, and the discrete measurements $\tilde{s}[i]$.}
\label{fig:MainModel}
\end{figure*}

\section{Signal Model}\label{sec:SignalModel}

In this section, we adapt the general framework of \cite{Unser2011_P1} to the continuous-time stochastic model studied in this paper. 
We follow the same notational conventions and write the input argument of the continuous-time signals/processes inside parenthesis (e.g., $s(\cdot)$) while we employ brackets (e.g., $s[\cdot]$) for discrete-time ones. Moreover, the \emph{tilde} diacritic is used to indicate the noisy signal. Typically, $\tilde{s}[\cdot]$ represents discrete noisy samples.

In Figure \ref{fig:MainModel}, we give a sketch of the model. The two main parts are the continuous-time innovation process and the linear operators. The process $s(\cdot)$ is generated by applying the shaping operator $\LI$ on the innovation process $w$. It can be whitened back by the inverse operator $\L$.
(Since the whitening operator is of greater importance, it is represented by $\L$ while $\LI$ refers to the shaping operator.) Furthermore, the discrete observations $\tilde{s}[\cdot]$ are formed by the noisy measurements of $s(\cdot)$. 

The innovation process and the linear operators have distinct implications on the resultant process $s$. Our model is able to handle general innovation processes that may or may not induce sparsity/compressibility. The distinction between these two cases is identified by a function $f(\omega)$ that is called the L\'evy exponent, as will be discussed in Section \ref{subsec:WhiteNoise}. The sparsity/compressibility of $s$ and, consequently, of the measurements $\tilde{s}$, is inherited from the innovations and is observed in a transform domain. This domain is tied to the operator $\L$. In this paper, we deal with operators that we represent by all-pole differential systems, tuned by acting upon the poles.

Although the model in Figure \ref{fig:MainModel} is rather classical for Gaussian innovations, the investigation of non-Gaussian innovations is nontrivial. While the transition from Gaussian to non-Gaussian necessitates the reinvestigation of every definition and result, it provides us with a more general class of stochastic processes which includes compressible/sparse signals.


\subsection{Innovation Process}\label{subsec:WhiteNoise}

Of all white processes, the Gaussian innovation is undoubtedly the one that has been investigated most thoroughly. However, it represents only a tiny fraction of the large family of white processes, which is best explored by using Gelfand's theory of generalized random processes.
In his approach, unlike with the conventional point-wise definition, the stochastic process is characterized through inner products with test functions.
 For this purpose, one first chooses a function space $\FS$ of test functions (e.g., the Schwartz class $\mathcal{S}$ of smooth and rapidly decaying functions). Then, one considers the random variable given by the inner product $\langle w,\varphi \rangle$, where $w$ represents the innovation process and $\varphi\in \FS$ \cite{Gelfandbook1964}.

\begin{defn}\label{defn:Innovation}
A stochastic process is called an innovation process if
\begin{enumerate}
\item it is stationary, \emph{i.e.}, the random variables $\langle w,\varphi_1 \rangle$ and $\langle w,\varphi_2\rangle$ are identically distributed, provided $\varphi_2$ is a shifted version of $\varphi_1$, and

\item it is white in the sense that the random variables $\langle w,\varphi_1 \rangle$ and $\langle w,\varphi_2\rangle$ are independent, provided $\varphi_1,\varphi_2\in\FS$ are non-overlapping test functions  (i.e., $\varphi_1\varphi_2\equiv 0$).
\end{enumerate}
\end{defn}
 
 The characteristic form of $w(\cdot)$ is defined as
\begin{eqnarray}\label{eq:CPPcharform}
\forall\,\varphi\in \FS:~~ \CH_w(\varphi)= \EV\big\{\e^{-\j\langle w,\varphi\rangle}\big\},
\end{eqnarray}
where $\EV\{\cdot\}$ represents the expected-value operator.
The characteristic form is a powerful tool for investigating the properties of random processes. For instance, it allows one to easily infer the probability density function of the random variable $\langle w,\varphi \rangle$, or the joint densities of $\langle w,\varphi_1 \rangle,\dots,\langle w,\varphi_n\rangle$. Further details regarding characteristic forms can be found in Appendix \ref{sec:AppCharForm}.

The key point in Gelfand's theory is to consider the form
\begin{eqnarray}\label{eq:GelfandGeneric}
\CH_w(\varphi)= \mathrm{exp}\left(\int_{\mathbb{R}} f\big(\varphi(x)\big)\d x\right).
\end{eqnarray}
and to provide the necessary and sufficient conditions on $f(\omega)$ (the L\'evy exponent) for $w$ to define a generalized innovation process over $\mathcal{S}^{'}$ (dual of $\mathcal{S}$). The class of admissible L\'evy exponents is characterized by the L\'evy–-Khintchine representation theorem \cite{Sato_book_1994,Steutel_book_2003} as
\ifthenelse{\equal{\Ncol}{two}}{
	\begin{eqnarray}\label{eq:LevyFunction_Gen}
	f(\omega)&=&\j \bbOne\omega-\frac{\bbTwo^2}{2}\omega^2\nonumber\\
	&+&\int_{\mathbb{R}\setminus\{0\}}\left(\e^{\j a\omega}-1-\j\omega a \bfOne_{]-1,1[}(a)\right) v(a)\,\d a,
	\end{eqnarray}
}{ 
	\begin{eqnarray}\label{eq:LevyFunction_Gen}
	f(\omega)=\j \bbOne\omega-\frac{\bbTwo^2}{2}\omega^2 +\int_{\mathbb{R}\setminus\{0\}}\left(\e^{\j a\omega}-1-\j\omega a \bfOne_{]-1,1[}(a)\right) v(a)\,\d a,
	\end{eqnarray}
}
where $\bfOne_{\mathcal{B}}(a)=1$ for $a\in\mathcal{B}$ and $0$ otherwise, and $v(\cdot)$ (the L\'evy density) is a real-valued density function that satisfies
\begin{eqnarray}\label{eq:LevyDensity_Constr}
\int_{\mathbb{R}\setminus\{0\}} \min(1,a^2) v(a)\,\d a<\infty.
\end{eqnarray}

In this paper, we  consider only symmetric real-valued L\'evy  exponents  (i.e., $\bbOne=0$ and $v(a)=v(-a)$). Thus, the general form of (\ref{eq:LevyFunction_Gen}) is reduced to
\begin{eqnarray}\label{eq:LevyFunction_Sym}
f(\omega)=-\frac{\bbTwo^2}{2}\omega^2 +\int_{\mathbb{R}\setminus\{0\}}\left(\cos(a\omega)-1\right) v(a)\,\d a.
\end{eqnarray}

Next, we discuss three particular cases of (\ref{eq:LevyFunction_Sym}) which are of special interest  in this paper.


\subsubsection{Gaussian Innovation}\label{subsubsec:Gaussian}
The choice $v\equiv 0$ turns (\ref{eq:LevyFunction_Sym}) into
\begin{eqnarray}
f_{\rm G}(\omega)=-\frac{\bbTwo^2}{2}\omega^2,
\end{eqnarray}
which implies
\begin{eqnarray}\label{eq:GaussianCharForm}
\CH_{w_{_{\rm G}}}(\varphi)= \e^{- \frac{\bbTwo^2}{2}\|\varphi\|_2^2}.
\end{eqnarray}
This shows that the random variable $\langle w_{_{\rm G}},\varphi\rangle$ has a zero-mean Gaussian distribution with  variance $\bbTwo^2\|\varphi\|_2^2$.


\subsubsection{Impulsive Poisson}\label{subsubsec:CompPoisson}
Let $\bbTwo=0$ and $v(a)=\lambda p_a(a)$, where $p_a$ is a symmetric probability density function. The corresponding white process is known as the impulsive Poisson innovation. By substitution in (\ref{eq:LevyFunction_Sym}), we obtain
\ifthenelse{\equal{\Ncol}{two}}{
	\begin{eqnarray}
	f_{\rm \IP}(\omega) &=& \lambda\int_{\mathbb{R}\setminus\{0\}}\left( \cos(a\omega)-1\right) p_a(a)\,\d a \nonumber\\
	&=& \lambda\big(\hat{p}_a(\omega)-1\big),
	\end{eqnarray}
}{
	\begin{eqnarray}
	f_{\rm \IP}(\omega)=\lambda\int_{\mathbb{R}\setminus\{0\}}\left( \cos(a\omega)-1\right) p_a(a)\,\d a =\lambda\big(\hat{p}_a(\omega)-1\big),
	\end{eqnarray}
}
where $\hat{p}_a$ denotes the Fourier transform of $p_a$. Let $\bfOne_{[0,1]}$ represents the test function that takes $1$ on $[0,1]$ and $0$ otherwise. Thus, if $X=\langle w_{\rm\IP},\bfOne_{[0,1]}\rangle$, then for the pdf of $X$ we know that (see Appendix \ref{sec:AppCharForm})
\ifthenelse{\equal{\Ncol}{two}}{
	\begin{eqnarray}\label{eq:CompPoisson_delta}
	p_X(x)&=&\mathcal{F}^{-1}_{\omega}\bigg\{\e^{\lambda (\hat{p}_a(\omega)-1)} \bigg\}(x) \nonumber\\
	&=& \e^{-\lambda}\mathcal{F}^{-1}_{\omega}\bigg\{\sum_{i=0}^{\infty}\frac{\big(\lambda\hat{p}_a(\omega)\big)^i}{i!} 		\bigg\}(x)\nonumber\\
	&=& \e^{-\lambda}\delta(x)+\sum_{i=1}^{\infty}\frac{\e^{-\lambda} \lambda^i}{i!}\big(\underbrace{p_a*\dots*p_a}_{i\,	\mathrm{times}}\big)(x).
	\end{eqnarray}
}{ 
	\begin{eqnarray}\label{eq:CompPoisson_delta}
	p_X(x) &=& \mathcal{F}^{-1}_{\omega}\bigg\{\e^{\lambda (\hat{p}_a(\omega)-1)} \bigg\}(x) = \e^{-\lambda}\mathcal{F}^{-1}_{\omega}\bigg\{\sum_{i=0}^{\infty}\frac{\big(\lambda\hat{p}_a(\omega)\big)^i}{i!} \bigg\}(x) \nonumber\\
	&=&  \e^{-\lambda}\delta(x)+\sum_{i=1}^{\infty}\frac{\e^{-\lambda} \lambda^i}{i!}\big(\underbrace{p_a*\dots*p_a}_{i\,	\mathrm{times}}\big)(x).
	\end{eqnarray}
}
It is not hard to check (see Appendix II in \cite{Unser2011} for a proof) that this distribution matches the one that we obtain by defining 
\begin{eqnarray}\label{eq:ImpulsiveDef}
w(x)=\sum_{k\in\mathbb{Z}}a_k\delta(x-x_k),
\end{eqnarray}
where $\delta(\cdot)$ stands for the Dirac distribution, $\big\{x_k\big\}_{k\in\mathbb{Z}}\subset \mathbb{R}$ is a sequence of random Poisson points with parameter $\lambda$, and $\{a_k\}_{k\in\mathbb{Z}}$ is an independent and identically distributed (i.i.d.) sequence with probability density $p_a$ independent of $\{x_k\}$. The sequence $\{x_k\}_{k\in\mathbb{Z}}$ is a Poisson point random sequence with parameter $\lambda$ if, for all real values $a<b<c<d$, the random variables $N_1=\big|\{x_k\}\cap [a,b]\big|$ and $N_2=\big|\{x_k\}\cap [c,d]\big|$ are independent and $N_1$ (or $N_2$) follows the Poisson distribution with mean $\lambda(b-a)$ (or $\lambda (d-c)$), which can be written as
\begin{eqnarray}
\textrm{Prob}\{N_1=n\}=\frac{\e^{-\lambda(b-a)}\big(\lambda(b-a)\big)^n}{n!}.
 \end{eqnarray}
 In \cite{Unser2011}, this type of innovation is introduced as a potential candidate for sparse processes, since all the inner products have a mass probability at $x=0$.


\subsubsection{Symmetric $\SO$-Stable}\label{subsubsec:AlphaStable}
The stable laws are probability density functions that are closed under convolution. More precisely, the pdf of a random variable $X$ is said to be stable if, for two independent and identical copies of $X$, namely, $X_1,X_2$, and for each pair of scalars $0\leq c_1,c_2$, there exists $0\leq c$ such that $c_1X_1+c_2X_2$ has the same pdf as $cX$. For stable laws, it is known that $c^{\SO}=c_1^{\SO}+c_2^{\SO}$ for some $0<\SO\leq 2$ \cite{Samorodnitsky_book_1994}; this is the reason why the law is indexed with $\SO$.  An $\SO$-stable law which corresponds to a symmetric pdf is called symmetric $\SO$-stable.
 It is possible to define symmetric $\SO$-stable white processes for $0<\SO<2$ by considering $\bbTwo=0$ and $v(a)=\frac{c_{\SO}}{|a|^{1+\SO}}$, where $0<c_{\SO}$. From (\ref{eq:LevyFunction_Sym}), we get

\ifthenelse{\equal{\Ncol}{two}}{
	\begin{small}
	\begin{eqnarray}
	f_{\rm S}(\omega)&=&c_{\SO}\int_{\mathbb{R}\setminus\{0\}}\frac{ \cos(a\omega)-1 }{|a|^{\SO+1}}\,\d a =-2c_{\alpha}\int_{\mathbb{R}}\frac{\sin^2\big( \frac{a\omega}{2}\big)}{|a|^{\SO+1}}\,\d a \nonumber\\
	&=& -2c_{\SO}|\omega|^{\SO}\int_{\mathbb{R}}\frac{\sin^2\big( \frac{x}{2}\big)}{|x|^{\SO+1}}\,\d x =-\bar{c}_{\SO}|\omega|^{\SO},
	\end{eqnarray}
	\end{small}
}{
	\begin{eqnarray}
	f_{\rm S}(\omega) &=& c_{\SO}\int_{\mathbb{R}\setminus\{0\}}\frac{  \cos(a\omega)-1 }{|a|^{\SO+1}}\,\d a =-2c_{\alpha}\int_{\mathbb{R}}\frac{\sin^2\big( \frac{a\omega}{2}\big)}{|a|^{\SO+1}}\,\d a \nonumber\\ 
	&=& -2c_{\SO}|\omega|^{\SO}\int_{\mathbb{R}}\frac{\sin^2\big( \frac{x}{2}\big)}{|x|^{\SO+1}}\,\d x = -\bar{c}_{\SO}|\omega|^{\SO},
	\end{eqnarray}
}
where $\bar{c}_{\SO}$ is a positive constant. This yields
\begin{eqnarray}
\CH_{w_{_{\rm S}}}(\varphi)=\e^{-\bar{c}_{\alpha}\|\varphi\|_{\SO}^{\SO}},
\end{eqnarray}
which confirms that every random variable of the form $\langle w_{_{\rm S}},\varphi\rangle$ has a symmetric $\SO$-stable distribution \cite{Samorodnitsky_book_1994}. The fat-tailed distributions including $\SO$-stables for $0<\SO<2$ are known to generate compressible sequences \cite{Amini2011}. Meanwhile, the Gaussian distributions are also stable laws that correspond to the extreme value $\SO=2$ and have classical and well-known properties that differ fundamentally from non-Gaussian laws.

The key message of this section is that the innovation process is uniquely determined by its L\'evy exponent $f(\omega)$. We shall explain in Section \ref{subsec:Sparsity} how $f(\omega)$ affects the sparsity and compressibility properties of the process $s$.


\subsection{Linear Operators}\label{subsec:LinearOperator}

The second important component of the model is the shaping operator (the inverse of the whitening operator $\L$) that determines the correlation structure of the process. For the generalized stochastic definition of $s$ in Figure \ref{fig:MainModel}, we expect to have
\begin{eqnarray}\label{eq:sVSw}
\langle s,\varphi \rangle=\langle \LI w, \mathnormal{\varphi} \rangle= \langle  w, \LI^{*} \varphi \rangle,
\end{eqnarray}
where $\LI^{*}$ represents the adjoint operator of $\LI$. It shows that $\LI^{*} \varphi$ ought to define a valid test function for the equalities in (\ref{eq:sVSw}) to remain valid. In turn, this sets constraints on $\LI$. The simplest choice for $\LI$ would be that of an operator which forms a continuous map from $\mathcal{S}$ into itself, but the class of such operators is not rich enough to cover the desired models in this paper. For this reason, we take advantage of a result in \cite{Unser2011_P1} that extends the choice of shaping operators to those $\LI$ operators for which $\LI^{*}$ forms a continuous mapping from $\mathcal{S}$ into $L_p$ for some $1\leq p$.


\subsubsection{Valid Inverse Operator $\LI$}\label{subsubsec:ShapingOp}
 In the sequel, we first explain the general requirements on the inverse of a given whitening operator $\L$. Then, we focus on a special class of operators $\L$ and study the implications for the associated shaping operators in more details.

We assume $\L$ to be a given whitening operator, which may or may not be uniquely invertible. The minimum requirement on the shaping operator $\LI$ is that it should form a right-inverse of $\L$ (i.e., $\L\LI=\I$, where $\I$ is the identity operator). Furthermore, since the adjoint operator is required in (\ref{eq:sVSw}), $\LI$ needs to be linear. This implies the existence of a kernel $h(x,\tau)$ such that
\begin{eqnarray}\label{eq:LinearLI}
\LI w(x) = \int_{\R} h(x,\tau) w(\tau)\d \tau.
\end{eqnarray}
Linear shift-invariant shaping operators are special cases that correspond to $h(x,\tau)=h(x-\tau)$. However, some of the $\LI$ operators considered in this paper are not shift-invariant. 

We require the kernel $h$ to satisfy the following three conditions:
\begin{itemize}
\item[(i)\phantom{ii}] $\L h(x,\tau) = \delta(x-\tau)$, where $\L$ acts on the parameter $x$ and $\delta$ is the Dirac function,
\item[(ii)\phantom{i}] $h(x,\tau)=0$, for $\tau>\max(0,x)$,
\item[(iii)] $(1+|\tau|^{p-1})\int_{\R}\frac{h(x,\tau)}{1+|x|^p}\d x$ is bounded for all $p\geq 1$.
\end{itemize}
Condition (i) is equivalent to $\L\LI=\I$, while (iii) is a sufficient condition studied in \cite{Unser2011_P1} to establish the continuity of the mapping $\LI:\mathcal{S}\mapsto L_p$, for all $p$. Condition (ii) is a constraint that we impose in this paper to simplify the statistical analysis. For $x\geq 0$, its implication is that the random variable $s(x)=\LI w(x)$ is fully determined by $w(\tau)$ with $\tau\leq x$, or, equivalently, it is independent of $w(\tau)$ for $\tau> x$.

From now on, we focus on differential operators $\L$ of the form $\sum_{i=0}^n \ARC_i\D^i$, where $\D$ is the first-order derivative ($\frac{\d}{\d x}$), $\D^0$ is the identity operator ($\I$),  and $\ARC_i$ are constants. 
 With Gaussian innovations, these operators generate the autoregressive processes. An equivalent representation of $\L$, which helps us in the analysis, is its decomposition into first-order factors as $\L=\ARC_n\prod_{i=1}^n (\D-r_i\I)$. 
The scalars $r_i$ are the roots of the characteristic polynomial and correspond to the poles of the inverse linear system.
Here, we assume that all the poles are in the left half-plane $\Re r_i\leq 0$. This assumption helps us associate the operator $\L$ to a suitable kernel $h$, as shown in Appendix \ref{sec:AppShapingKernel}.

Every differential operator $\L$ has a unique causal Green function $\rho_{\L}$ \cite{Unser2007p1}. The linear shift-invariant system defined by $h(x,\tau)=\rho_{\L}(x-\tau)$ satisfies conditions (i)-(ii). If all the poles strictly lie in the left half-plane (i.e., $\Re r_i<0$), due to absolute integrability of $\rho_{\L}$ (stability of the system), $h(x,\tau)$ satisfies condition (iii) as well. The definition of $\LI$ given  through the kernels in Appendix \ref{sec:AppShapingKernel} achieves both linearity and stability, while loosing shift-invariance when $\L$ contains poles on the imaginary axis. 
It is worth mentioning that the application of two different right-inverses of $\L$ on a given input produces results that differ only by an exponential polynomial that is in the null space of $\L$.


\subsubsection{Discretization}\label{subsubsec:Discretization}
Apart from qualifying as whitening operators, differential operators have other appealing properties such as the existence of finite-length discrete counterparts. To explain this concept, let us first consider the first-order continuous-time derivative operator $\D$ that is associated with the finite-difference filter $H(z)=1-z^{-1}$. This discrete counterpart is of finite length (FIR filter). Further, for any right inverse of $\D$ such as $\D^{-1}$, the system $\D_{d,T} \D^{-1}$ is shift invariant and its impulse response is compactly supported. Here, $\D_{d,T}$ is the discretized operator corresponding to the sampling period $T$ with impulse response $\big(\delta(\cdot)-\delta(\cdot-T)\big)$. It should be emphasized that the discrete counterpart $H(z)$ is a discrete-domain operator, while the discretized operator acts on continuous-domain signals. It is easy to check that this impulse response coincides with the causal B-spline of degree 0 ($\bfOne_{[0,1[}$).
 In general, the discrete counterpart of $\L=\ARC_n \prod_{i=1}^n (\D-r_i\I)$ is defined through its factors. Each $\D-r_i\I$ is associated with its discrete counterpart $H_i(z)=1-\e^{r_i}z^{-1}$ and a discretized operator given by the impulse response $\delta(\cdot)-\e^{r_iT}\delta(\cdot-T)$. The convolution of $n$ such impulse responses gives rise to the impulse response of $\L_{d,T}$ (up to the scaling factor $\ARC_n$), which is the discretized operator of $\L$ for the sampling period $T$. By expanding the convolution, we obtain the form $\sum_{i=0}^{n}d_T[k]\delta(\cdot-kT)$ for the impulse response of $\L_{d,T}$. It is now evident that $\L_{d,T}$ corresponds to an FIR filter of length $(n+1)$ represented by $\{d_T[k]\}_{k=0}^{n}$ with $d_T[0]\neq 0$. Results in spline theory confirm that, for any right inverse $\L^{-1}$ of $\L$, the operator $\L_{d,T}\LI$ is shift invariant and the support of its impulse response is contained in $[0,nT)$ \cite{Unser2005ES1}. The compactly supported impulse response of $\L_{d,T}\LI$, which we denote by $\SP(\cdot)$, is usually referred to as the $\L$-spline. We define the generalized finite differences by
\ifthenelse{\equal{\Ncol}{two}}{
	\begin{eqnarray}\label{eq:dTdefinition}
	u_T(x)&=&\big(\SP*\,w\big)(x) =(\L_{d,T}\LI\, w)(x)\nonumber\\
	&=&(\L_{d,T}\, s)(x)=\sum_{k=0}^{n}d_T[k]s(x-kT) .
	\end{eqnarray}
}{
	\begin{eqnarray}\label{eq:dTdefinition}
	u_T(x) = \big(\SP*\,w\big)(x) =(\L_{d,T}\LI\, w)(x) = (\L_{d,T}\, s)(x)=\sum_{k=0}^{n}d_T[k]s(x-kT) .
	\end{eqnarray}
}

We show in Figures \ref{fig:StationaryModel} (a), (b)  the definitions of $\SP(x)$ and $u_T(x)$, respectively.

\ifthenelse{\equal{\Ncol}{two}}{
	\begin{figure}[tb]
		\centering
		\begin{minipage}[b]{\linewidth}
  			\centering
  			\includegraphics[width=7cm]{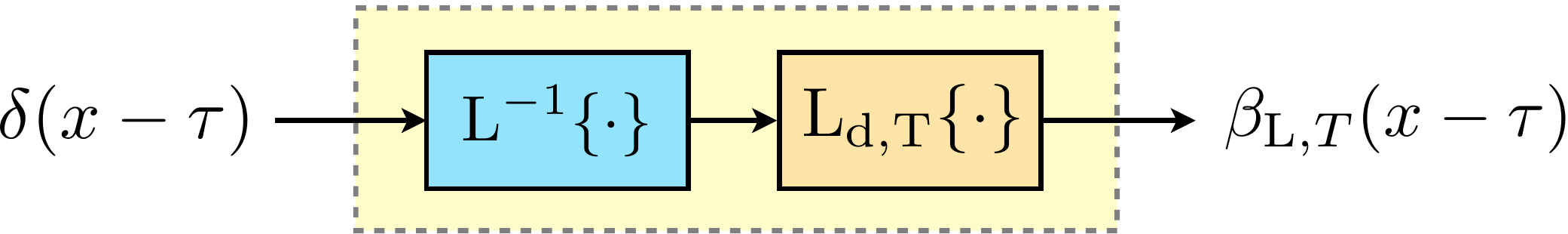}
  			\centerline{(a)}\medskip
		\end{minipage}
		\hfill
		\begin{minipage}[b]{\linewidth}
  			\centering
  			\includegraphics[width=5cm]{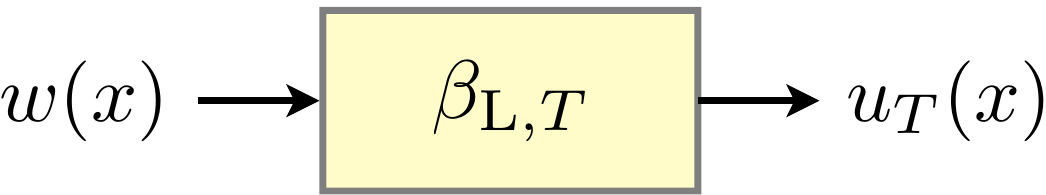}
  			\centerline{(b)}\medskip
		\end{minipage}
		\caption{(a) Linear shift-invariant operator $\L_{d,T} \LI$ and its impulse response $\SP$ ($\L$-spline). (b) Definition of the auxiliary signal $u_T(x)$.}
		\label{fig:StationaryModel}
	\end{figure}
}{
	\begin{figure*}[tb]
		\centering
		\begin{minipage}[b]{0.54\linewidth}
  			\centering
  			\includegraphics[width=7cm]{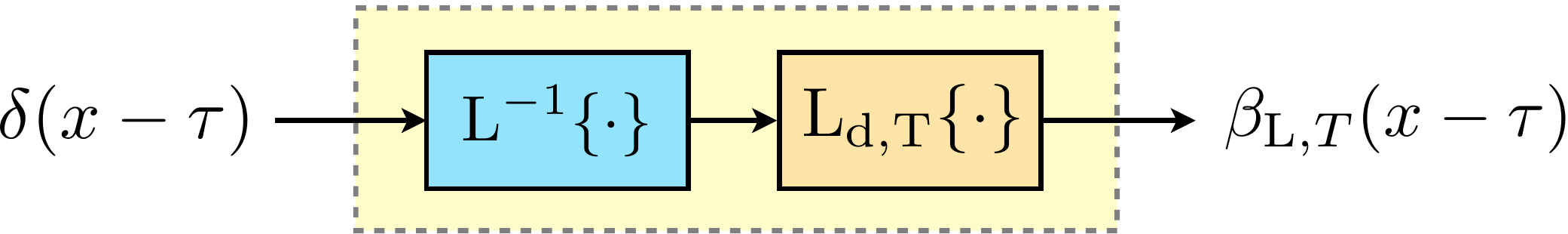}
  			\centerline{(a)}\medskip
		\end{minipage}
		\begin{minipage}[b]{0.39\linewidth}
  			\centering
  			\includegraphics[width=5cm]{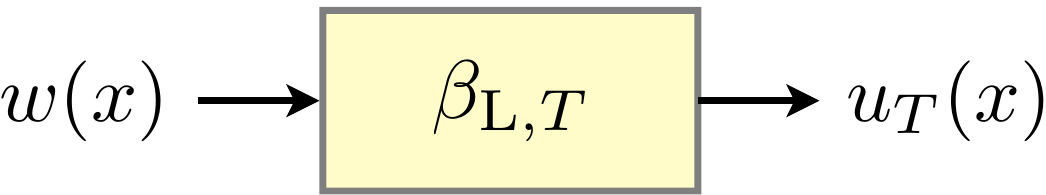}
  			\centerline{(b)}\medskip
		\end{minipage}
		\caption{(a) Linear shift-invariant operator $\L_{d,T} \LI$ and its impulse response $\SP$ ($\L$-spline). (b) Definition of the auxiliary signal $u_T(x)$.}
		\label{fig:StationaryModel}
	\end{figure*}
}


\subsection{Sparsity/Compressibility}\label{subsec:Sparsity}

The innovation process can be thought of as a concatenation of independent atoms. A consequence of this independence is that the process contains no redundancies. Therefore, it is incompressible under unique representation constraint. 

In our framework, the role of the shaping operator $\LI$ is to generate a specific correlation structure in $s$ by mixing the atoms of $w$. Conversely, the whitening operator $\L$ undoes the mixing and returns an incompressible set of data, in the sense that it maximally compresses the data. For a discretization of $s$ corresponding to a sampling period $T$, the operator $\L_{d,T}$ mimics the role of $\L$. It efficiently uncouples the sequence of samples and produces the generalized differences $u_T$, where each term depends only on a finite number of other terms. 
Thus, the role of $\L_{d,T}$ can be compared to that of converting discrete-time signals into their transform domain representation.
As we explain now, the sparsity/compressibility properties of $u_T$ are closely related to the L\'evy exponent $f$ of $w$.

The concept is best explained by focusing on a special class known as L\'evy processes that correspond to $\SP(x)=\bfOne_{[0,T[}(x)$ (see Section \ref{sec:Levy} for more details). By using (\ref{eq:GelfandGeneric}) and (\ref{eq:ProbFourierCharForm}), we can check that the characteristic function of the random variable $u_T$ is given by $\e^{Tf(\omega)}$. 
When the L\'evy function $f$ is generated through a nonzero density $v(a)$ that is absolutely integrable (i.e., impulsive Poisson), the pdf of $u_T$ associated with $\e^{Tf(\omega)}$ necessarily contains a mass at the origin \cite{Steutel_book_2003} (Theorems 4.18 and 4.20). This is interpreted as sparsity when considering a finite number of measurements. 

It is shown in \cite{Amini2011}, \cite{Gribonval2012} that the compressibility of the measurements depends on the tail of their pdf. In simple terms, if the pdf decays at most inverse-polynomially, then, it is compressible in some corresponding $L_p$ norm. The interesting point is that the inverse-polynomial decay of an infinite-divisible pdf is equivalent to the inverse-polynomial decay of its L\'evy density $v(\cdot)$ with the same order \cite{Steutel_book_2003} (Theorem 7.3). Therefore, an innovation process with a fat-tailed L\'evy density results in processes with compressible measurements. This indicates that the slower the decay rate of $v(\cdot)$, the more compressible  the measurements of $s$.


\subsection{Summary of the Parameters of the Model}\label{subsec:Guide}

We now briefly review the degrees of freedom in the model and how the dependent variables are determined. The innovation process is uniquely determined by the L\'evy triplet $(\bbOne,\bbTwo,v)$. The sparsity/compressibility of the process can be determined through the L\'evy density $v$ (see Section \ref{subsec:Sparsity}). The values $n$ and $\{\frac{\ARC_i}{\ARC_n}\}_{i=0}^{n-1}$, or, equivalently, the poles $\{r_i\}_{i=1}^n$ of the system, serve as the free parameters for the whitening operator $\L$. As explained in Section \ref{subsubsec:Discretization}, the taps of the FIR filter $d_T[i]$ are determined by the poles.


\section{Prior Distribution}\label{subsec:aPriorDist}

To derive statistical estimation methods such as MAP and MMSE, we need the a priori distributions. In this section, by using the generalized differences in (\ref{eq:dTdefinition}), we obtain the prior distribution and factorize it efficiently. This factorization is fundamental, since it makes the implementation of the MMSE and MAP estimators tractable.

The general problem studied  in this paper is to estimate $s(x)$ at $\big\{x=i\frac{m}{m_s}\big\}_{i=0}^{m_s}$ for arbitrary $m_s\in\mathbb{N}^{*}$ (values of the continuous process $s$ at a uniform sampling grid with $T=\frac{m}{m_s}$), given a finite number $(m+1)$ of noisy/noiseless measurements $\{\tilde{s}[k]\}_{k=0}^m$ of $s(x)$ at the integers. Although we are aware that this is not equivalent to estimating the continuous process, by increasing $m_s$ we are able to make the estimation grid as fine as desired. For piecewise-continuous signals, the limit process of refining the grid can give us access to the properties of the continuous domain. 

To simplify the notations let us define
\begin{eqnarray}
\left\{\begin{array}{lll}
s_{T}[i] & = & s(x)|_{x=iT}, \\
u_{T}[i] & = & u_T(x)|_{x=iT},
\end{array}\right. 
\end{eqnarray}
where $u_T(x)$ is defined in (\ref{eq:dTdefinition}).
Our goal is to derive the joint distribution of $s_{T}[i]$ (a priori distribution). However, the $s_{T}[i]$ are in general pairwise-dependent, which makes it difficult to deal with the joint distribution in high dimensions. This corresponds to a large number of samples. 
Meanwhile, as will be shown in Lemma \ref{lemma:indep}, the sequence $\{u_{T}[i]\}_i$  forms a Markov chain of order $(n-1)$ that helps in factorizing the joint probability distributions, whereas $\{s_{T}[i]\}_i$ does not.
The leading idea of this work is then that each $u_{T}[i]$ depends on a finite number of $u_{T}[j]$, $j\neq i$. It then becomes  much simpler to derive the joint distribution of $\{u_{T}[i]\}_i$ and link it to that of $\{s_{T}[i]\}_i$. Lemma \ref{lemma:indep} helps us to factorize the joint pdf of $\{u_{T}[i]\}_i$.

\begin{lemma}\label{lemma:indep}
For $N\geq n$ and $i\geq 0$, where $n$ is the differential order of  $\L$, 
\begin{enumerate}
\item the random variables $\{u_{T}[i]\}_i$ are identically distributed,

\item the sequence $\{u_{T}[i]\}_i$ is a Markov chain of order $n-1$, and

\item the sample $u_{T}[i+N]$ is statistically independent of $u_{T}[i]$ and $s_{T}[i]$.
\end{enumerate}
\end{lemma}

\textbf{Proof.} First note that
\begin{eqnarray}\label{eq:uTWN}
u_{T}[i]=\big(\SP*w\big)(x)|_{x=iT}=\langle w\,,\,\SP(iT-\cdot)\rangle.
\end{eqnarray}
Since $\SP(iT-\cdot)$ functions are shifts of the same function for various $i$ and $w$ is stationary (Definition \ref{defn:Innovation}), $\{u_{T}[i]\}_i$ are identically distributed.

Recalling that $\SP(\cdot)$ is supported on $[0,nT)$, we know that $\SP(iT-\cdot)$ and $\SP\big((i+N)T-\cdot\big)$ have no common support for $N\geq n$. Thus, due to the whiteness of $w$ \emph{c.f.} Definition \ref{defn:Innovation}), the random variables $\langle w\,,\,\SP(iT-\cdot)\rangle$ and $\langle w\,,\,\SP\big((i+N)T-\cdot\big)\rangle$ are independent. Consequently, we can write
\ifthenelse{\equal{\Ncol}{two}}{
	\begin{eqnarray}
	&p_u\big( u_T[i+n]~\big|~u_T[i+n-1],u_T[i+n-2],\dots\big)& \nonumber\\
	&=p_u\big(u_T[i+n]~\big|~u_T[i+n-1],\dots,u_T[i+1]\big),&
	\end{eqnarray}
}{
	\begin{small}
	\begin{eqnarray}
	p_u\big( u_T[i+n]~\big|~u_T[i+n-1],u_T[i+n-2],\dots\big) = p_u\big(u_T[i+n]~\big|~u_T[i+n-1],\dots,u_T[i+1]\big),
	\end{eqnarray}
	\end{small}
}
which confirms that the sequence $\{u_{T}[i]\}_i$ forms a Markov chain of order $(n-1)$. Note that the choice of $N\geq n$ is due to the support of $\SP$. If the support of $\SP$ was infinite, it would be impossible to find $j$ such that $u_{T}[i]$ and $u_{T}[j]$ were independent in the strict sense.

To prove the second independence property, we recall that
\begin{eqnarray}
s_T[i] = s(x)\big|_{x=iT}= \int_{\R} h(iT,\tau)w(\tau)\d\tau = \langle w , h(iT,\cdot)\rangle.
\end{eqnarray}
Condition (ii) in Section \ref{subsubsec:ShapingOp} implies that $h(iT,\tau)=0$ for $\tau>\max(0,iT)$. Hence, $h(iT,\tau)$ and $\SP\big((i+N)T-\cdot\big)$ have disjoint supports. Again due to whiteness of $w$, this implies that $u_T[i+N]$ and $s_T[i]$ are independent. $\hspace{\stretch{1}}\blacksquare$

We now exploit the properties of $u_T[i]$ to obtain the a priori distribution of $s_T[i]$. Theorem \ref{theo:s2dprob}, which is proved in Appendix \ref{sec:AppTheoFactorization} summarizes the main result of Section \ref{subsec:aPriorDist}.

\begin{theo}\label{theo:s2dprob}
Using the conventions of Lemma \ref{lemma:indep}, for $k\geq 2n-1$ we have
\ifthenelse{\equal{\Ncol}{two}}{
	\begin{eqnarray}\label{eq:theoJointS}
	&&p_s\big(s_T[k],\dots,s_T[0]\big)=\nonumber\\
	&&~~~~\prod_{\theta=2n-1}^{k} \big| d_T[0]\big|\;p_{u}\Big(u_T[\theta]~\Big|~\big\{u_T[\theta-i]\big\}_{i=1}^{n-1}\Big)\nonumber\\
	&&~~~~\times p_{s}\big(s_T[2n-2],\dots,s_T[0]\big).
	\end{eqnarray}
}{
	\begin{eqnarray}\label{eq:theoJointS}
	p_s\big(s_T[k],\dots,s_T[0]\big) &=& \prod_{\theta=2n-1}^{k} \big| d_T[0]\big|\;p_{u}\Big(u_T[\theta]~\Big|~\big\{u_T[\theta-i]\big\}_{i=1}^{n-1}\Big)\nonumber\\
	&&~~\times p_{s}\big(s_T[2n-2],\dots,s_T[0]\big).
	\end{eqnarray}
}
\end{theo}

In the definition of $\LI$ proposed in Section \ref{subsec:LinearOperator}, except when all the poles are strictly included in the left half-plane, the operator $\LI$ fails to be shift-invariant. Consequently, neither $s(x)$ nor $s_T[i]$ are stationary. An interesting consequence of Theorem \ref{theo:s2dprob} is that it relates the probability distribution of the non-stationary process $s_T[i]$ to that of the stationary process $u_T[i]$ plus a minimal set of transient terms.

Next, we show in Theorem \ref{theo:Cond_d_CharForm} how the conditional probability of $u_T[i]$ can be obtained from a characteristic form. To maintain the flow of the paper, the proof is postponed to Appendix \ref{sec:AppTheoPDF}.

\begin{theo}\label{theo:Cond_d_CharForm}
The probability density function of $u_T[\AI]$ conditioned on $(n-1)$ previous $u_T[i]$ is given by
\ifthenelse{\equal{\Ncol}{two}}{
	\begin{eqnarray}
	& p_{u}\Big(u_T[\AI]~\Big|~\big\{u_T[\AI-i]\big\}_{i=1}^{n-1}\Big)=&\nonumber\\
	& \frac{ \mathcal{F}^{-1}_{\{\omega_{i}\}}\Big\{  \e^{ I_{w,\SP}(\omega_0,\dots,\omega_{n-1}) } \Big\} \big(\{u_T[\AI-i]\}_{i=0}^{n-1}\big) }{ \mathcal{F}^{-1}_{\{\omega_{i}\}}\Big\{ \e^{ I_{w,\SP}(0,\omega_1,\dots,\omega_{n-1}) }\Big\} 	\big(\{u_T[\AI-i]\}_{i=1}^{n-1}\big)  }& ,
	\end{eqnarray}
}{
	\begin{eqnarray}
	 p_{u}\Big(u_T[\AI]~\Big|~\big\{u_T[\AI-i]\big\}_{i=1}^{n-1}\Big) = \frac{ \mathcal{F}^{-1}_{\{\omega_{i}\}}\Big\{  \e^{ I_{w,\SP}(\omega_0,\dots,\omega_{n-1}) } \Big\} \big(\{u_T[\AI-i]\}_{i=0}^{n-1}\big) }{ \mathcal{F}^{-1}_{\{\omega_{i}\}}\Big\{ \e^{ I_{w,\SP}(0,\omega_1,\dots,\omega_{n-1}) }\Big\} 	\big(\{u_T[\AI-i]\}_{i=1}^{n-1}\big)  } ,
	\end{eqnarray}
}
where
\ifthenelse{\equal{\Ncol}{two}}{
	\begin{eqnarray}\label{eq:Ifunc}
	& I_{w,\SP}(\omega_0,\dots,\omega_{n-1}) \,:= \nonumber\\
	& \int_{\mathbb{R}}f_w\Big(\sum_{i=0}^{n-1}\omega_i\SP(x-iT)\Big)\d x.
	\end{eqnarray}
}{
	\begin{eqnarray}\label{eq:Ifunc}
	I_{w,\SP}(\omega_0,\dots,\omega_{n-1}) \,:= \int_{\mathbb{R}}f_w\Big(\sum_{i=0}^{n-1}\omega_i\SP(x-iT)\Big)\d x.
	\end{eqnarray}
}
\end{theo}


\section{Signal Estimation}\label{sec:Estimation}
 MMSE and MAP estimation are two common statistical paradigms for signal recovery. Since the optimal MMSE estimator is rarely realizable in practice, MAP is often used as the next best thing.
 In this section, in addition to applying the two methods to the proposed class of signals, we settle the question of knowing when MAP is a good approximation of MMSE.

For the estimation purpose it is convenient to assume that the sampling instances associated with $\tilde{s}[i]$ are included in the uniform sampling grid for which we want to estimate the values of $s$. In other words, we assume that $T=\frac{m}{m_s}=\frac{1}{n_T}$, where $T$ is the sampling period in the definition of $s_T[\cdot]$ and $n_T$ is a positive integer. This assumption does impose no lower bound on the resolution of the grid because we can set $T$ arbitrarily close to zero by increasing $n_T$. 

To simplify mathematical formulations, we use vectorial notations. We indicate the vector of noisy/noiseless measurements $\{\tilde{s}[i]\}_{i=0}^{m}$ by $\tilde{\mathbf{s}}$. The vector $\mathbf{s}_T$ stands for the hypothetical realization $\{s_T[k]\}_{k=0}^{mn_T} $ of the process on the considered grid, and $\mathbf{s}_{T,n_T}$ denotes the subset $\{s_T[in_T]\}_{i=0}^{m}$ that corresponds to the points at which we have a sample. Finally, we represent the vector of estimated values $\{\hat{s}_T[k]\}_{k=0}^{mn_T}$ by $\hat{\mathbf{s}}_T$.


\subsection{MMSE Denoising}\label{subsec:MMSEEst}
It is very common to evaluate the quality of an estimation method by means of the mean-square error, or SNR. In this regard, the best estimator, known as MMSE, is obtained by evaluating the posterior mean, or $\hat{\mathbf{s}}_T=\EV\{\mathbf{s}_T \,|\,\tilde{\mathbf{s}}\}$. 

For Gaussian processes, this expectation is easy to obtain, since it is equivalent to the best linear estimator \cite{Blu2007p2}. However, there are only a few non-Gaussian cases where an explicit closed form of this expectation is known. In particular, if the additive noise is white and Gaussian (no restriction on the distribution of the signal) and pure denoising is desired ($T=1$), then the MMSE estimator can be reformulated as
\begin{eqnarray}\label{eq:MMSE_orig}
\hat{\mathbf{s}}_{ \rm MMSE}=\tilde{\mathbf{s}}+\sigma_n^2\mathbf{\nabla} \log p_{\tilde{s}}(\mathbf{x})\big|_{\mathbf{x}=\tilde{\mathbf{s}}}\;,
\end{eqnarray}
where $\hat{\mathbf{s}}_{\rm MMSE}$ stands for $\hat{\mathbf{s}}_{T, {\rm MMSE}}$ with $T=1$, and $\sigma_n^2$ is the variance of the noise \cite{Stein1981,Raphan2007,Luisier2007}. Note that the pdf $p_{\tilde{s}}$, which is the result of convolving the a priori distribution $p_s$ with the Gaussian pdf of the noise, is both continuous and differentiable.


\subsection{MAP Estimator}\label{subsec:MAPEst}
Searching for the MAP estimator amounts to finding the maximizer of the distribution $p(\mathbf{s}_T \;|\;\tilde{\mathbf{s}})$. It is commonplace to reformulate this conditional probability in terms of the a priori distribution. 

The additive discrete noise $\tilde{n}$ is white and Gaussian with the variance $\sigma_n^2$. Thus,
\begin{eqnarray}\label{eq:APosterioriProb}
p\big(\mathbf{s}_{T}\,\big|\, \tilde{\mathbf{s}}\big) = p\big(\tilde{\mathbf{s}} \,\big|\, \mathbf{s}_{T}\big)  \frac{p_s(\mathbf{s}_{T} )}{p_{\tilde{s}}( \tilde{\mathbf{s}} )} = \frac{\e^{\frac{-1}{2\sigma_n^2} \|\tilde{\mathbf{s}}-\mathbf{s}_{T,n_T}\|_2^2 }}{(2\pi\sigma_n^2)^{\frac{m+1}{2}}} \frac{p_s(\mathbf{s}_{T} )}{p_{\tilde{s}}( \tilde{\mathbf{s}} )}.
\end{eqnarray}
In MAP estimation, we are looking for a vector $\mathbf{s}_T$  that maximizes the conditional a posteriori probability, so that (\ref{eq:APosterioriProb}) leads to
\ifthenelse{\equal{\Ncol}{two}}{
	\begin{eqnarray}\label{eq:MAPraw}
	\hat{\mathbf{s}}_{T,{\rm MAP}} &=& \mathrm{arg}\max_{ \mathbf{s}_T }~p\big(\mathbf{s}_T \,\big|\, \tilde{\mathbf{s}}\big) \nonumber\\
	&=& \mathrm{arg}\max_{ \mathbf{s}_T }~\frac{\e^{\frac{-1}{2\sigma_n^2} \|\tilde{\mathbf{s}}-\mathbf{s}_{T,n_T}\|_2^2 } \, p_s(\mathbf{s}_T )}{(2\pi\sigma_n^2)^{\frac{m+1}{2}} \, p_{\tilde{s}}( \tilde{\mathbf{s}} ) }   \nonumber\\
	&=& \mathrm{arg}\max_{ \mathbf{s}_T }~\e^{\frac{-1}{2\sigma_n^2} \|\tilde{\mathbf{s}}-\mathbf{s}_{T,n_T}\|_2^2 } p_s(\mathbf{s}_T ).
	\end{eqnarray}
}{
	\begin{eqnarray}\label{eq:MAPraw}
	\hat{\mathbf{s}}_{T,{\rm MAP}} =\mathrm{arg}\max_{ \mathbf{s}_T }~p\big(\mathbf{s}_T \big| \tilde{\mathbf{s}}\big) = \mathrm{arg}\max_{ \mathbf{s}_T }~\frac{\e^{-\frac{\|\tilde{\mathbf{s}}-\mathbf{s}_{T,n_T}\|_2^2}{2\sigma_n^2}  } \, p_s(\mathbf{s}_T )}{(2\pi\sigma_n^2)^{\frac{m+1}{2}} \, p_{\tilde{s}}( \tilde{\mathbf{s}} ) }   = \mathrm{arg}\max_{ \mathbf{s}_T }~\e^{-\frac{\|\tilde{\mathbf{s}}-\mathbf{s}_{T,n_T}\|_2^2}{2\sigma_n^2}  } p_s(\mathbf{s}_T ).
	\end{eqnarray}
}
The last equality is due to the fact that neither $(2\pi\sigma_n^2)^{\frac{m+1}{2}}$ nor $p_{\tilde{s}}( \tilde{\mathbf{s}} )$ depend on the choice of $\mathbf{s}_T$. Therefore, they play no role in the maximization.

If the pdf of $\mathbf{s}_T$ is bounded, the cost function (\ref{eq:MAPraw}) can be replaced with its logarithm without changing the maximizer. The equivalent logarithmic maximization problem is given by 
\begin{eqnarray}\label{eq:firstMAP}
\hat{\mathbf{s}}_{T,{\rm MAP}} = \mathrm{arg}\min_{ \mathbf{s}_T } \|\mathbf{s}_{T,n_T} - \tilde{\mathbf{s}}\|_2^2 - 2\sigma_n^2 \log p_s(\mathbf{s}_T). 
\end{eqnarray}
By using the pdf factorization provided by Theorem \ref{theo:s2dprob}, (\ref{eq:firstMAP}) is further simplified to 
\ifthenelse{\equal{\Ncol}{two}}{
	\begin{eqnarray}\label{eq:secondMAP}
	\hat{\mathbf{s}}_{T,{\rm MAP}} &=& \mathrm{arg}\min_{ \mathbf{s}_T } \bigg\{ \|\mathbf{s}_{T,n_T} - \tilde{\mathbf{s}}\|_2^2 \nonumber\\
	&& \,- 2\sigma_n^2 \sum_{k=2n-1}^{mn_T}\log p_u\big(u_T[k]\big| \big\{u_T[k-i]\big\}_{i=1}^{n-1}\big) \nonumber\\
	&&\,  -2\sigma_n^2\log p_s\big(s_T[2n-2],\dots,s_T[0]\big) \bigg\},
	\end{eqnarray}
}{
	\begin{eqnarray}\label{eq:secondMAP}
	\hat{\mathbf{s}}_{T,{\rm MAP}} &=& \mathrm{arg}\min_{ \mathbf{s}_T } \bigg\{ \|\mathbf{s}_{T,n_T} - \tilde{\mathbf{s}}\|_2^2 - 2\sigma_n^2 \sum_{k=2n-1}^{mn_T}\log p_u\big(u_T[k]\big| \big\{u_T[k-i]\big\}_{i=1}^{n-1}\big) \nonumber\\
	&& \phantom{\mathrm{arg}\min_{ \mathbf{s}_T } \bigg\{ } \,  -2\sigma_n^2\log p_s\big(s_T[2n-2],\dots,s_T[0]\big) \bigg\},
	\end{eqnarray}
}
where each $u_T[i]$ is provided by the linear combination of the elements in $\mathbf{s}_T$ found in (\ref{eq:dTdefinition}). 
If we denote the term $\left(-\log p_u\big(u_T[k]\big| \big\{u_T[k-i]\big\}_{i=1}^{n-1}\big)\right)$ by $\Psi_T\big(u_T[k],\dots,u_T[k-n+1]\big)$, the MAP estimation  becomes equivalent to the minimization problem
\ifthenelse{\equal{\Ncol}{two}}{
	\begin{eqnarray}\label{eq:MainMAP}
	\hat{\mathbf{s}}_{T,{\rm MAP}} &=& \mathrm{arg}\min_{\mathbf{s}_T}~ \bigg\{ \|\mathbf{s}_{T,n_T}-\tilde{\mathbf{s}}\|_2^2 \nonumber\\
	&& \,  +\lambda \sum_{k=2n-1}^{mn_T}\Psi_T\big(u_T[k],\dots,u_T[k-n+1]\big) \nonumber\\
	&& \,  -\lambda \, \log p_s\big(s_T[2n-2],\dots,s_T[0]\big) \bigg\},
	\end{eqnarray}
	where $\lambda =2\sigma_n^2$. The interesting aspect is that the MAP estimator has the same form as (\ref{eq:PenaltyDenoising}) where the sparsity-promoting term $\Psi_T$ in the cost function is determined by both $\LI$ and the distribution of the innovation.
The well-known and successful TV regularizer corresponds to the special case where $\Psi_T(\cdot)$ is the univariate function $|\cdot|$ and the FIR filter $d_T[\cdot]$ is the finite-difference operator. In Appendix \ref{sec:AppTV_MAP}, we show the existence of an innovation process for which the MAP estimation coincides with the TV regularization.
}{
	\begin{eqnarray}\label{eq:MainMAP}
	\hat{\mathbf{s}}_{T,{\rm MAP}} &=& \mathrm{arg}\min_{\mathbf{s}_T}~ \bigg\{ \|\mathbf{s}_{T,n_T}-\tilde{\mathbf{s}}\|_2^2 +\lambda \sum_{k=2n-1}^{mn_T}\Psi_T\big(u_T[k],\dots,u_T[k-n+1]\big) \nonumber\\
	&& \phantom{\mathrm{arg}\min_{\mathbf{s}_T}~ \bigg\{} \,  -\lambda\, \log p_s\big(s_T[2n-2],\dots,s_T[0]\big) \bigg\},
	\end{eqnarray} 
	where $\lambda =2\sigma_n^2$. The interesting aspect is that the MAP estimator has the same form as (\ref{eq:PenaltyDenoising}) where the sparsity-promoting term $\Psi_T$ in the cost function is determined by both $\LI$ and the distribution of the innovation.
The well-known and successful TV regularizer corresponds to the special case where $\Psi_T(\cdot)$ is the univariate function $|\cdot|$ and the FIR filter $d_T[\cdot]$ is the finite-difference operator. In Appendix \ref{sec:AppTV_MAP}, we show the existence of an innovation process for which the MAP estimation coincides with the TV regularization.
}


\subsection{MAP \emph{vs} MMSE}\label{subsec:MAP_MMSE}

To have a rough comparison of MAP and MMSE, it is beneficial to reformulate the MMSE estimator in (\ref{eq:MMSE_orig}) as a variational problem similar to (\ref{eq:MainMAP}), thereby, expressing the MMSE solution as the minimizer of a cost function that consists of a quadratic term and a sparsity-promoting penalty term. In fact, for sparse priors, it is shown in \cite{Gribonval2011a} that the minimizer of a cost function involving the $\ell_1$-norm penalty term approximates the MMSE estimator more accurately than the commonly considered MAP estimator. Here, we propose a different interpretation without going into technical details. From (\ref{eq:MMSE_orig}), it is clear that
\begin{eqnarray}\label{eq:MMSE_reformulated}
\hat{\mathbf{s}}_{\rm MMSE} =\tilde{\mathbf{s}}+\sigma_n^2\mathbf{\nabla}\log p_{\tilde{s}}\big(\hat{\mathbf{s}}_{\rm MMSE}-\mathbf{b}_{\tilde{s}}\big),
\end{eqnarray}
where $\mathbf{b}_{\tilde{s}} = \sigma_n^2\mathbf{\nabla} \log p_{\tilde{s}}\big(\tilde{\mathbf{s}}\big)$.
We can check that $\hat{\mathbf{s}}_{\rm MMSE}$ in (\ref{eq:MMSE_reformulated}) sets the gradient of the cost $J(\mathbf{s})=\|\mathbf{s}-\tilde{\mathbf{s}}\|_2^2-2\sigma_n^2 \log p_{\tilde{s}}(\mathbf{s}-\mathbf{b}_{\tilde{s}})$ to zero. It suggests that
\begin{eqnarray}\label{eq:MMSE_minimizer}
	\hat{\mathbf{s}}_{\rm MMSE} = \text{arg}\min_{\mathbf{s}} \|\mathbf{s}-\tilde{\mathbf{s}}\|_2^2-2\sigma_n^2 \log p_{\tilde{s}}(\mathbf{s}-\mathbf{b}_{\tilde{s}} ). 
\end{eqnarray}
which is similar to (\ref{eq:firstMAP}). The latter result is only valid when the cost function has a unique minimizer. Similarly to \cite{Gribonval2011a}, it is possible to show that, under some mild conditions, this constraint is fulfilled. Nevertheless, for the qualitative comparison of MAP and MMSE, we only focus on the local properties of the  cost functions that are involved. The main distinction between the cost functions in (\ref{eq:MMSE_minimizer}) and (\ref{eq:firstMAP}) is the required pdf. For MAP, we need $p_s$, which was shown to be factorizable by the introduction of generalized finite differences. For MMSE, we require $p_{\tilde{s}}$. Recall that $p_{\tilde{s}}$ is the result of convolving $p_s$ with a Gaussian pdf. Thus, irrespective of the discontinuities of $p_s$, the function $p_{\tilde{s}}$ is smooth. However, the latter is no longer separable, which complicates the minimization task. The other difference is the offset term $\mathbf{b}_{\tilde{s}}$ in the MMSE cost function. For heavy-tail innovations such as $\SO$-stables, the convolution by the Gaussian pdf of the noise does not greatly affect $p_s$. In such cases, $p_{\tilde{s}}$ can be approximated by $p_s$ fairly well, indicating that the MAP  estimator suffers from a bias ($\mathbf{b}_{\tilde{s}}$). The effect of convolving $p_s$ with a Gaussian pdf becomes more evident as $p_s$ decays faster. In the extreme case where $p_s$ is Gaussian, $p_{\tilde{s}}$ is also Gaussian (convolution of two Gaussians) with a different mean (which introduces another type of bias). The fact that MAP and MMSE estimators are equivalent for Gaussian innovations indicates that the two biases act in opposite directions and cancel out each other. In summary, for super-exponentially decaying innovations, MAP seems to be consistent with MMSE. For heavy-tail innovations, however, the MAP estimator is a biased version of the MMSE, where the effect of the bias is observable at high noise powers. The scenario in which MAP diverges most from MMSE might be the exponentially decaying innovations, where we have both a mismatch and a bias in the cost function, as will be confirmed in the experimental part of the paper.


\section{Example: L\'evy Process}\label{sec:Levy}

To demonstrate the results and implications of the theory, we consider L\'evy processes as special cases of the model in Figure \ref{fig:MainModel}. L\'evy processes are roughly defined as processes with stationary and independent increments that start from zero. The processes are compatible with our model by setting $\L=\frac{\d}{\d x}$ (i.e., $n=1$ and $r_1=0$), $\LI=\int_{0}^x$, or $h(x,\tau)=\bfOne_{[0,\infty[}(x-\tau)-\bfOne_{[0,\infty[}(-\tau)$ which imposes the boundary condition $s(0)=\int_0^0 w(\tau) d\tau=0$.
The corresponding discrete FIR filter has the two taps $d_T[0]=1$ and $d_T[1]=-1$. The impulse response of $\L_{d,T}\LI$ is given by
\begin{eqnarray}
\SP(x)=u(x)-u(x-T)=\left\{\begin{array}{ll}
1, & 0\leq x< T\\
0, & \textrm{otherwise}.
\end{array}\right. 
\end{eqnarray}


\subsection{MMSE Interpolation}
A classical problem is to interpolate the signal using noiseless samples. This corresponds to the estimation of $s_T[\AI]$ where $n_T\nmid \AI$  ($n_T$ does not divide $\AI$) by assuming $\tilde{s}[i]=s_T[in_T]$ ($0\leq i\leq m$). Although there is no noise in this scenario, we can still employ the MMSE criterion to estimate $s_T[\AI]$. 
We show that the MMSE interpolator of a L\'evy process is the simple linear interpolator, irrespective of the type of innovation. To prove this, we assume $l\,n_T<\AI<(l+1)n_T$ and rewrite $s_T[\AI]$ as 
\begin{eqnarray}
s_T[\AI] &=& s_T[ln_T] + \sum_{k=ln_T+1}^{\AI} \underbrace{ s_T[k]-s_T[k-1] }_{ u_T[k] }.
\end{eqnarray}
This enables us to compute
\ifthenelse{\equal{\Ncol}{two}}{
	\begin{eqnarray}\label{eq:MMSE_interp1}
	\hat{s}_T[\AI]&=&\EV\big\{s_T[\AI]~\big|~\{s_T[in_T]\}_{i=0}^{m}\big\}\nonumber\\
	&=&s_T[ln_T]+\EV\Big\{ \sum_{k=ln_T+1}^{\AI} u_T[k] \Big| \{s_T[kn_T]\}_i\Big\} .
	\end{eqnarray}
}{
	\begin{eqnarray}\label{eq:MMSE_interp1}
	\hat{s}_T[\AI] = \EV\big\{s_T[\AI]~\big|~\{s_T[in_T]\}_{i=0}^{m}\big\} = s_T[ln_T]+\EV\Big\{ \sum_{k=ln_T+1}^{\AI} u_T[k] \Big| \{s_T[in_T]\}_k\Big\} .
	\end{eqnarray}
}
Since the mapping from the set $\{s_T[in_T]\}_i$ to $\{s_T[0]\}\cup\{u_1[i]=s_T[(i+1)n_T]-s_T[in_T]\}_i$ is one to one, the two sets can be used interchangeably for evaluating the conditional expectation. Thus,
\begin{align}\label{eq:MMSE_interp1_1}
	\hat{s}_T[\AI] = s_T[ln_T]+\sum_{k=ln_T+1}^{\AI} \EV\Big\{  u_T[k] \Big| \{s_T[0]\}\cup\{u_1[i]\}_i \Big\}.
\end{align}
According to Lemma \ref{lemma:indep}, $u_T[k]$ (for $k>0$) is independent of $s_T[0]$ and $u_T[k^{'}]$, where $k\neq k^{'}$. By rewriting $u_1[i]$ as $\sum_{k^{'}=in_T+1}^{(i+1)n_T}u_T[k^{'}]$, we conclude that $u_1[i]$ is independent of $u_T[k]$ unless $in_T+1\leq k\leq (i+1)n_T$. Hence,
\begin{eqnarray}\label{eq:MMSEinterp_bef_aux}
	\hat{s}_T[\AI]=s_T[ln_T]+\sum_{k=ln_T+1}^{\AI}\EV\Big\{u_T[k]~\Big|~u_1[l+1]\Big\}.
\end{eqnarray}
Since $u_1[l+1]=\sum_{i=ln_T+1}^{(l+1)n_T}u_T[i]$ and $\{u_T[i]\}_i$ is a sequence of i.i.d. random variables, the expected mean of $u_T[i]$  conditioned on $u_1[l+1]$ is the same for all $i$  with $ln_T+1\leq i\leq (l+1)n_T$, which yields
\ifthenelse{\equal{\Ncol}{two}}{
	\begin{eqnarray}\label{eq:MMSEinterp_aux}
	\EV\Big\{u_T[k]\Big|u_1[l+1]\Big\} &=& \frac{1}{n_T} \sum_{i=ln_T+1}^{(l+1)n_T}\EV\Big\{u_T[i]\Big|u_1[l+1]\Big\}\nonumber\\
	&=&\frac{1}{n_T} \EV\Big\{\sum_{i=ln_T+1}^{(l+1)n_T}u_T[i]\Big|u_1[l+1]\Big\} \nonumber\\
	&=&\frac{u_1[l+1]}{n_T}.
	\end{eqnarray}
}{
	\begin{eqnarray}\label{eq:MMSEinterp_aux}
	\EV\Big\{u_T[i]\Big|u_1[l+1]\Big\} &=& \frac{1}{n_T} \sum_{k=ln_T+1}^{(l+1)n_T}\EV\Big\{u_T[k]\Big|u_1[l+1]\Big\}\nonumber\\
	&=&\frac{1}{n_T} \EV\Big\{\sum_{k=ln_T+1}^{(l+1)n_T}u_T[k]\Big|u_1[l+1]\Big\} =\frac{u_1[l+1]}{n_T}.
	\end{eqnarray}
}
By applying (\ref{eq:MMSEinterp_aux}) to (\ref{eq:MMSEinterp_bef_aux}), we obtain
\ifthenelse{\equal{\Ncol}{two}}{
	\begin{eqnarray}\label{eq:MMSE_interp2}
	\hat{s}_T[\AI]&=&s_T[ln_T]+\frac{\AI-ln_T}{n_T} u_1[l+1]\nonumber\\
	&=& (1-\lambda_{\AI}) \,s_T[ln_T] + \lambda_{\AI} \, s_T[(l+1)n_T],
	\end{eqnarray}
}{
	\begin{eqnarray}\label{eq:MMSE_interp2}
	\hat{s}_T[\AI]=s_T[ln_T]+\frac{\AI-ln_T}{n_T} u_1[l+1]= (1-\lambda_{\AI}) \,s_T[ln_T] + \lambda_{\AI} \, s_T[(l+1)n_T],
	\end{eqnarray}
}
where $\lambda_{\AI}=\frac{\AI-ln_T}{n_T}$. Obviously, (\ref{eq:MMSE_interp2}) indicates a linear interpolation between the samples.


\subsection{MAP Denoising}\label{subsec:LevyMAPdenoising}

Since $n=1$, the finite differences $u_T[i]$ are independent. Therefore, the conditional probabilities involved in Theorem \ref{theo:s2dprob} can be replaced with the simple pdf values
\begin{eqnarray}
p_s\big(s_T[k],\dots,s_T[0]\big)=p_s\big(s_T[0]\big)\prod_{\theta=1}^kp_u\big(u_T[\theta]\big).
\end{eqnarray}
In addition, since $f_{w}(0)=0$, from Theorem \ref{theo:Cond_d_CharForm}, we have
\begin{eqnarray}\label{eq:Levy_pdf}
\left\{\begin{array}{ccl}
I_{w,\SP}( \omega ) & = & T f_w( \omega ) \\
p_u\big(u_T[\theta]\big) & = &\mathcal{F}^{-1}_{\omega}\big\{ \e^{Tf_w(\omega)} \big\}\big(u_T[\theta]\big) .
\end{array}\right. 
\end{eqnarray}

\begin{figure}
\centering
\includegraphics[width=8.5cm]{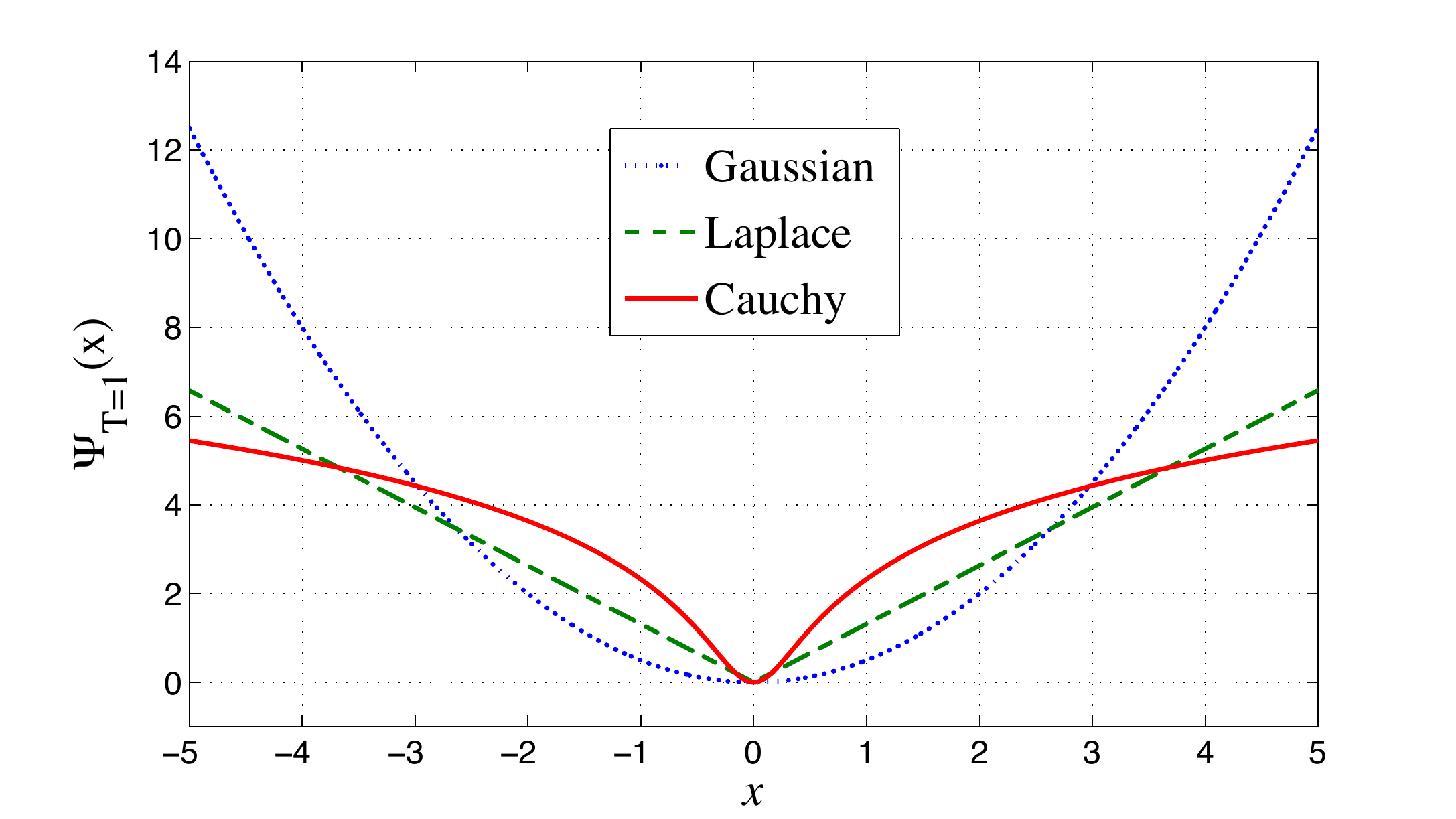}
\caption{The $\Psi_T(\cdot)=-\log p_u(\cdot)$ functions at $T=1$ for Gaussian, Laplace, and Cauchy distributions. The parameters of each pdf are tuned such that they all have the same entropy  and the curves are shifted to enforce them to pass through the origin.}
\label{fig:PsiFunctions}
\end{figure}

In the case of impulsive Poisson innovations, as shown in (\ref{eq:CompPoisson_delta}), the pdf of $u_T[i]$ has a single mass probability at $x=0$. Hence, the MAP estimator will choose $u_T[i]=0$ for all $i$, resulting in a constant signal. In other words, according to the MAP criterion and due to the boundary condition $s(0)=0$, the optimal estimate is nothing but the trivial all-zero function. For the other types of innovations where the pdf of the increments $u_T[i]$ is bounded, or, equivalently, when the L\'evy density $v(\cdot)$ is singular at the origin \cite{Steutel_book_2003}, one can reformulate the MAP estimation in the form of (\ref{eq:MainMAP}) as
\ifthenelse{\equal{\Ncol}{two}}{
	\begin{eqnarray}\label{eq:LevyMAPdenoiser}
	&\hat{s}_T[0]=0& ~~~~~~~~ \mathrm{and} \nonumber\\
	& \big\{\hat{s}_T[k]\big\}_{k=1}^{mn_T}=&  \mathrm{arg}\min_{s_T[k]}\Big\{\sum_{i=1}^{m}\big(\tilde{s}[i]-s_T[in_T]\big)^2\nonumber\\
	& \phantom{\big\{\hat{s}_T[k]\big\}_{k=1}^{mn_T}=}& + \lambda \Psi_T\big(s_T[1]\big) \nonumber\\
	& \phantom{\big\{\hat{s}_T[k]\big\}_{k=1}^{mn_T}=}& + \lambda \sum_{k=2}^{mn_T}\Psi_T\big(s_T[k]-s_T[k-1]\big)\Big\},
	\end{eqnarray}
}{
	\begin{eqnarray}\label{eq:LevyMAPdenoiser}
	&& \hat{s}_T[0]=0 ~~~~~~~~ \mathrm{and} \nonumber\\
	&&\big\{\hat{s}_T[k]\big\}_{k=1}^{mn_T}=\mathrm{arg}\min_{s_T[k]}\Big\{\sum_{i=1}^{m}\big(\tilde{s}[i]-s_T[in_T]\big)^2\nonumber+ 2\sigma_n^2\Psi_T\big(s_T[1]\big)  \nonumber\\
	&&~~~~~~~~~~~~~~~~~~  + 2\sigma_n^2\sum_{k=2}^{mn_T}\Psi_T\big(s_T[k]-s_T[k-1]\big)\Big\},
	\end{eqnarray}
}
where $\Psi_T(\cdot)=-\log p_u(\cdot)$ and $\lambda=2\sigma_n^2$. Because shifting the function $\Psi_T$ with a fixed additive scalar does not change the minimizer of (\ref{eq:LevyMAPdenoiser}), we can modify the function to pass through the origin  (i.e., $\Psi_T(0)=0$).
After having applied this modification, the function $\Psi_{T=1}$ presents itself as shown in Figure \ref{fig:PsiFunctions} for various innovation processes such as

\begin{enumerate}

\item Gaussian innovation: $\bbTwo=1$, and $v(a)\equiv 0$, which implies
\begin{eqnarray}
p_u(x)=\frac{\e^{-\frac{x^2}{2}}}{\sqrt{2\pi}}.
\end{eqnarray}

\item Laplace-type innovation: $\bbTwo=0$, $v(a)=\frac{\e^{-\sqrt{\frac{2\e}{\pi}}|a|}}{|a|}$, which implies (see Appendix \ref{sec:AppTV_MAP})
\begin{eqnarray}
p_u(x)=\sqrt{\frac{\e}{2\pi}} \e^{-\sqrt{\frac{2\e}{\pi}}|x|} .
\end{eqnarray}
The L\'evy process of this innovation is known as the \emph{variance gamma process} \cite{Madan1998}.

\item Cauchy innovation ($\SO$-stable with $\SO=1$): $\bbTwo=0$, $v(a)=\frac{\sqrt{\frac{\e}{8\pi^3}}}{a^2}$, which implies
\begin{eqnarray}
p_u(x)=\frac{\sqrt{\frac{8 \e}{\pi}}}{\e+8\pi x^2} .
\end{eqnarray}

\end{enumerate}

The parameters of the above innovations are set such that they all lead to the same entropy value $\frac{\log(2\pi\e)}{2}\approx 1.41$. The negative log-likelihoods of the first two innovation types resemble the $\ell_2$ and $\ell_1$ regularization terms. However, the curve of $\Psi_T$ for the Cauchy innovation shows a nonconvex log-type function.

\subsection{MMSE Denoising}\label{sec:MessagePassing}
As discussed in Section \ref{subsec:MAP_MMSE}, the MMSE estimator, either in the expectation form or as a minimization problem, is not separable with respect to the inputs. This is usually a critical restriction in high dimensions. Fortunately, due to the factorization of the joint a priori distribution, we can lift this restriction by employing the powerful message-passing algorithm. The method consists of representing the statistical dependencies between the parameters as a graph and finding the marginal distributions by iteratively passing the estimated pdfs along the edges \cite{Loeliger.etal2007}. The transmitted messages along the edges are also known as \emph{beliefs}, which give rise to the alternative name of \emph{belief propagation}. In general, the marginal distributions (beliefs) are continuous-domain objects. Hence, for computer simulations we need to discretize them.

In order to define a graphical model for a given joint probability distribution, we need to define two types of nodes: \emph{variable} nodes that represent the input arguments for the joint pdf and \emph{factor} nodes that portray each one of the terms in the factorization of the joint pdf. The edges in the graph are drawn only between nodes of different type and indicate the contribution of an input argument to a given factor. 

\ifthenelse{\equal{\Ncol}{two}}{
For the L\'evy process, we consider the joint conditional pdf $p\big(\{s_T[k]\}_k\;\big|\;\{\tilde{s}[i]\}_i\big)$ factorized as
}{
For the L\'evy process, the joint conditional pdf $p\big(\{s_T[k]\}_k\;\big|\;\{\tilde{s}[i]\}_i\big)$ is factorized as
}
\ifthenelse{\equal{\Ncol}{two}}{
	\begin{eqnarray}
	\label{MP:CondProba}
	& p\Big(\{s_{T}[k]\}_{k=1}^{m}\;\Big|\;\{\tilde{s}[i]\}_{i=1}^{m}\Big)& \nonumber\\
 	& = \frac{ \prod_{i=1}^{m} \mathcal{G}\left( \tilde{s}[i] - s_T[i]; \sigma_n^2 \right) \prod_{k=1}^{mn_T}p_u \left( s_T[k] - s_T[k-1] \right) }{Z}, &
	\end{eqnarray}
}{
	\begin{eqnarray}
	\label{MP:CondProba}
	p\Big(\{s_{T}[k]\}_{k=1}^{m}\;\Big|\;\{\tilde{s}[i]\}_{i=1}^{m}\Big)  = \frac{1}{Z} \prod_{i=1}^{m} \mathcal{G}\left( \tilde{s}[i] - s_T[i]; \sigma_n^2 \right) \prod_{k=1}^{mn_T}p_u \left( s_T[k] - s_T[k-1] \right),
	\end{eqnarray}
}
where $Z = p_{\tilde{s}}\left(\{\tilde{s}[i]\}_{i=1}^{m}\right)$ is a normalization constant that depends only on the noisy measurements and $\mathcal{G}$ is the Gaussian function defined as
\begin{equation}
\label{MP:Gaussian}
\mathcal{G}\left( x; \sigma^2 \right) = \frac{1}{\sigma\sqrt{2\pi}}\mathrm{e}^{-\frac{x^2}{2\sigma^2}}.
\end{equation}
Note that, by definition, we have $s_T[0] = 0$.

\begin{figure}[tb]
\centering
\includegraphics[width=8.5cm]{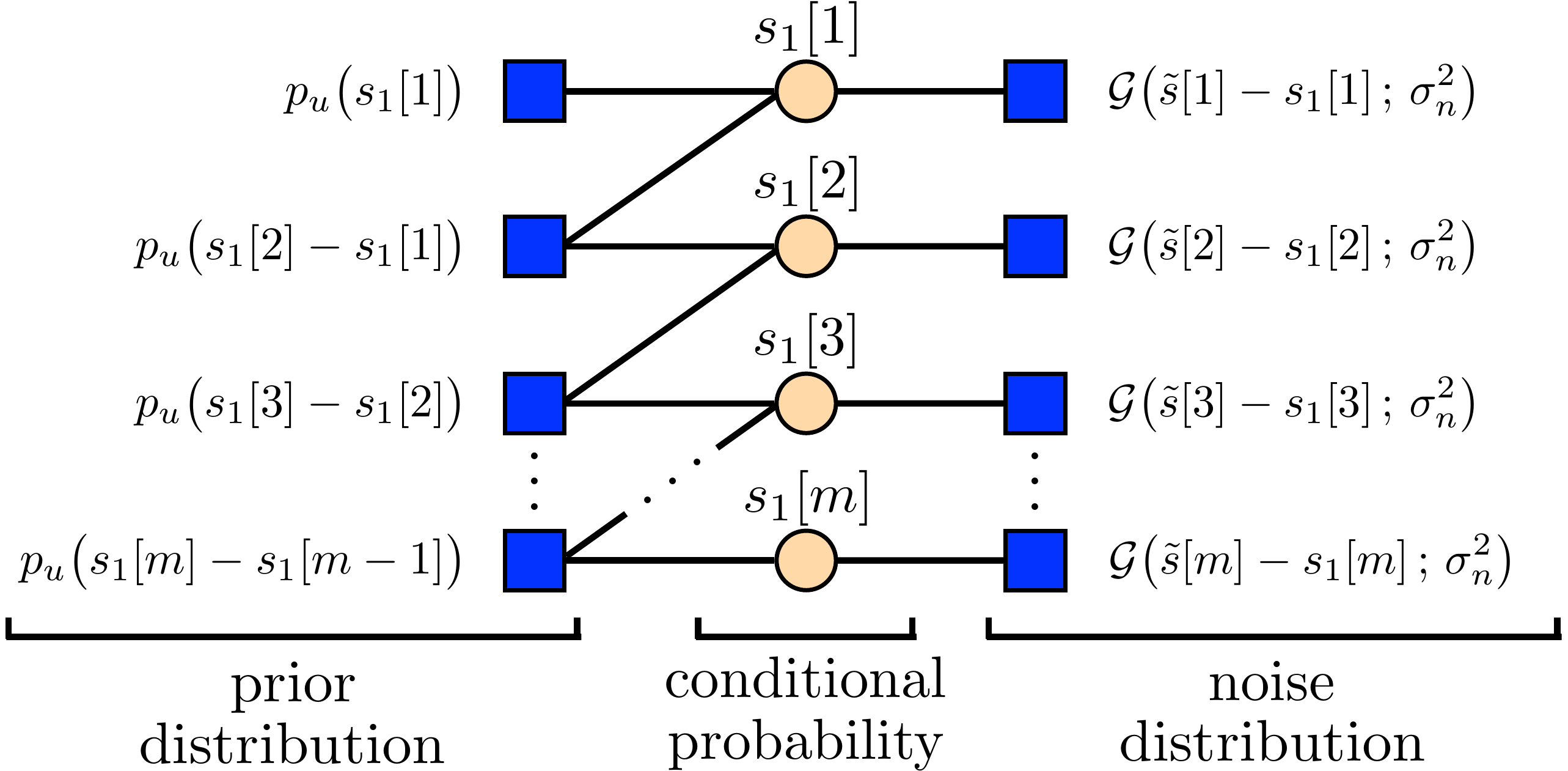}
\caption{Factor graph for the MMSE denoising of a L\'evy process. There are $m$ variable nodes (circles) and $2m$ factor nodes (squares).}
\label{MP:fig:factorGraph}
\end{figure}

For illustration purposes, we consider the special case of pure denoising corresponding to $T=1$. We give in Figure \ref{MP:fig:factorGraph} the bipartite graph $G = (V, F, E)$ associated to the joint pdf \eqref{MP:CondProba}. The variable nodes $V = \{1, \dots, m\}$ depicted in the middle of the graph stand for the input arguments $\{s_{T}[k]\}_{k=1}^{m}$. The factor nodes $F = \{1, \dots, 2m\}$ are placed at the right and left sides of the variable nodes depending on whether they represent the Gaussian factors or the $p_u(\cdot)$ factors, respectively. The set of edges $E = \{ (i, a) \in V \times F \}$ also indicates a participation of the variable nodes in the corresponding factor nodes. 

The message-passing algorithm consists of initializing the nodes of the graph with proper 1D functions and updating these functions through communications over the graph. It is desired that we eventually obtain the marginal pdf $p\big( s_1[k] \;\big|\;\{\tilde{s}[i]\}_{i=1}^{m} \big)$ on the $k$th variable node, which enables us to obtain the mean.
The details of the messages sent over the edges and updating rules are given in \cite{Kamilov2012conf}, \cite{Kamilov2012}.


\section{Simulation Results}\label{sec:SimulationResults}
For the experiments, we consider the denoising of L\'evy processes for various types of innovation, including those introduced in Section \ref{subsec:WhiteNoise} and the Laplace-type innovation discussed in Appendix \ref{sec:AppTV_MAP}. Among the heavy-tail $\SO$-stable innovations, we choose the Cauchy distribution corresponding to $\SO=1$. The four implemented denoising methods are
\begin{enumerate}

\item Linear minimum mean-square error (LMMSE) method or quadratic regularization (also known as smoothing spline \cite{Unser2005}) defined as
\begin{eqnarray}
\mathrm{arg}\min_{s[i]} \Big\{ \|\mathbf{s}-\tilde{\mathbf{s}}\|_{\ell_2}^2+\lambda \sum_{i=1}^m \big(s[i]-s[i-1]\big)^2\Big\},
\end{eqnarray}
where $\lambda$ should be optimized. For finding the optimum $\lambda$ for given innovation statistics and a given additive-noise variance, we search for the best $\lambda$ for each realization by comparing the results with the oracle estimator provided by the noiseless signal. Then, we average $\lambda$ over a number of realizations to obtain a unified and realization-independent value. This procedure is repeated each time the statistics (either the innovation or the additive noise) change. For Gaussian processes, the LMMSE method coincides with both the MAP and MMSE estimators.

\item Total-variation regularization represented as
\begin{eqnarray}
\mathrm{arg}\min_{s[i]} \Big\{ \|\mathbf{s}-\tilde{\mathbf{s}}\|_{\ell_2}^2+\lambda \sum_{i=1}^m \big|s[i]-s[i-1]\big|\Big\},
\end{eqnarray}
where $\lambda$ should be optimized. The optimization process for $\lambda$ is similar to the one explained for the LMMSE method.

\item Logarithmic (Log) regularization described by

\ifthenelse{\equal{\Ncol}{two}}{
	\begin{footnotesize}
	\begin{eqnarray}
	\mathrm{arg}\min_{s[i]} \Big\{ \|\mathbf{s}-\tilde{\mathbf{s}}\|_{\ell_2}^2+\lambda \sum_{i=1}^m \log\Big(1+\frac{(s[i]-s[i-1])^2}	{\epsilon^2}\Big)\Big\},
	\end{eqnarray}
	\end{footnotesize}
}{
	\begin{eqnarray}
	\mathrm{arg}\min_{s[i]} \Big\{ \|\mathbf{s}-\tilde{\mathbf{s}}\|_{\ell_2}^2+\lambda \sum_{i=1}^m \log\Big(1+\frac{(s[i]-s[i-1])^2}	{\epsilon^2}\Big)\Big\},
	\end{eqnarray}
}
where $\lambda$ should be optimized. The optimization process is similar to the one explained for the LMMSE method. In our experiments, we keep $\epsilon=1$ fixed throughout the minimization steps (e.g., in the gradient-descent iterations).
Unfortunately, Log is not necessarily convex, which might result in a nonconvex cost function. Hence, it is possible that gradient-descent methods get trapped in local minima rather than the desired global minimum. For heavy-tail innovations (e.g., $\SO$-stables), the Log regularizer is either the exact, or a very good approximation of, the MAP estimator. 

\item Minimum mean-square error denoiser which is implemented using the message-passing technique discussed in Section  \ref{sec:MessagePassing}.
\end{enumerate}

The experiments are conducted in MATLAB. We have developed a graphical user interface that facilitates the procedures of generating samples of the stochastic process and denoising them using MMSE or the variational techniques.

\ifthenelse{\equal{\Ncol}{two}}{
	\begin{figure}
	\centering
	\includegraphics[width=8.5cm]{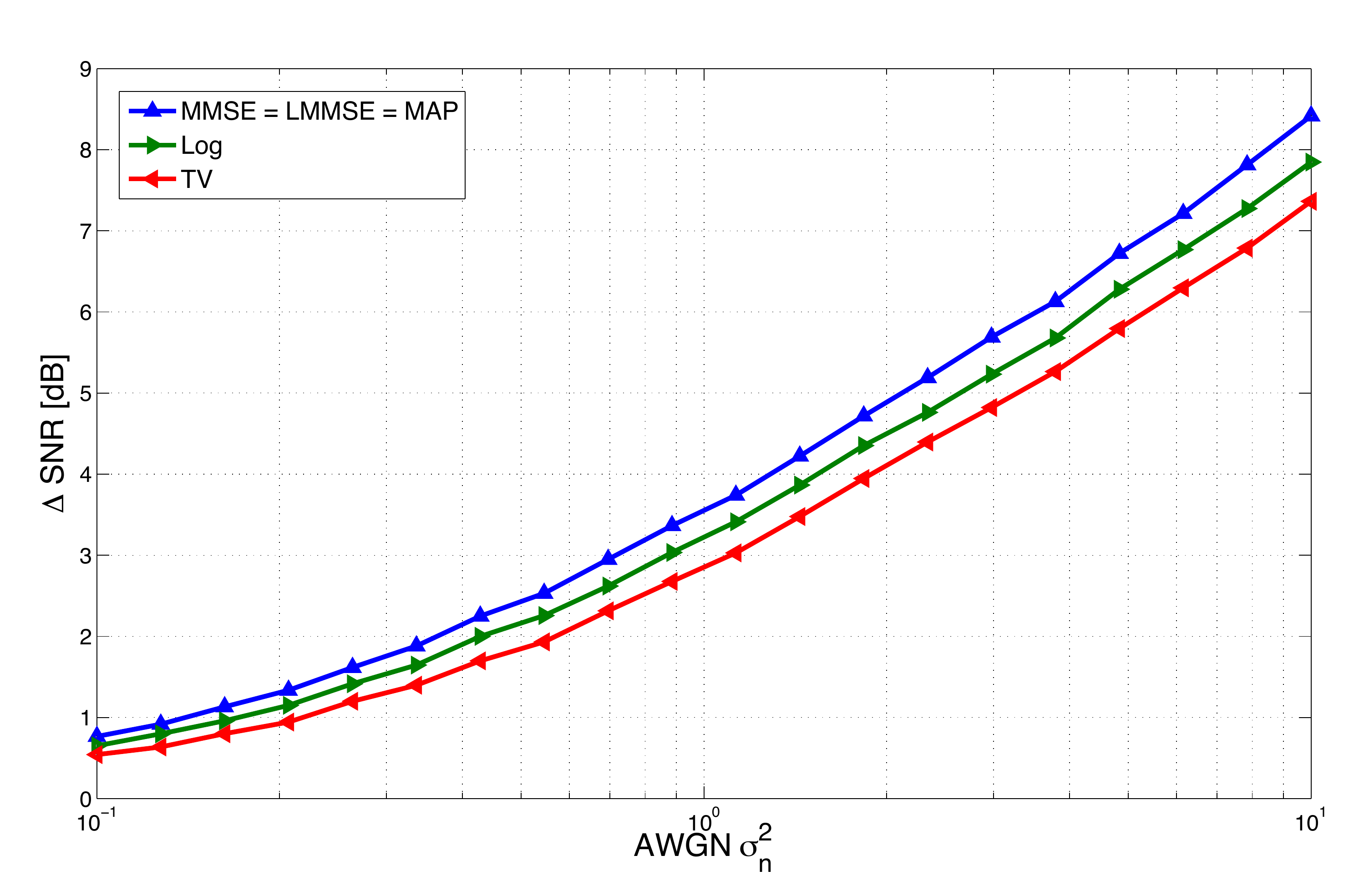}
	\caption{SNR improvement vs. variance of the additive noise for Gaussian innovations. The denoising methods are: MMSE estimator (which is equivalent to MAP and LMMSE estimators here), Log regularization, and TV regularization.}
	\label{fig:Gaussian_SNRimprovement}
	\end{figure}
}{
	\begin{figure}[t]
		\begin{minipage}[b]{0.5\linewidth}
			\centering
			\includegraphics[width=8cm]{Gaussian.pdf}
			\caption{SNR improvement vs. variance of the additive noise for Gaussian innovations. The denoising methods are: MMSE estimator (which is equivalent to MAP and LMMSE estimators here), Log regularization, and TV regularization.}
			\label{fig:Gaussian_SNRimprovement}
		\end{minipage}
		\hspace{0.5cm}
		\begin{minipage}[b]{0.5\linewidth}
			\centering
			\includegraphics[width=8cm]{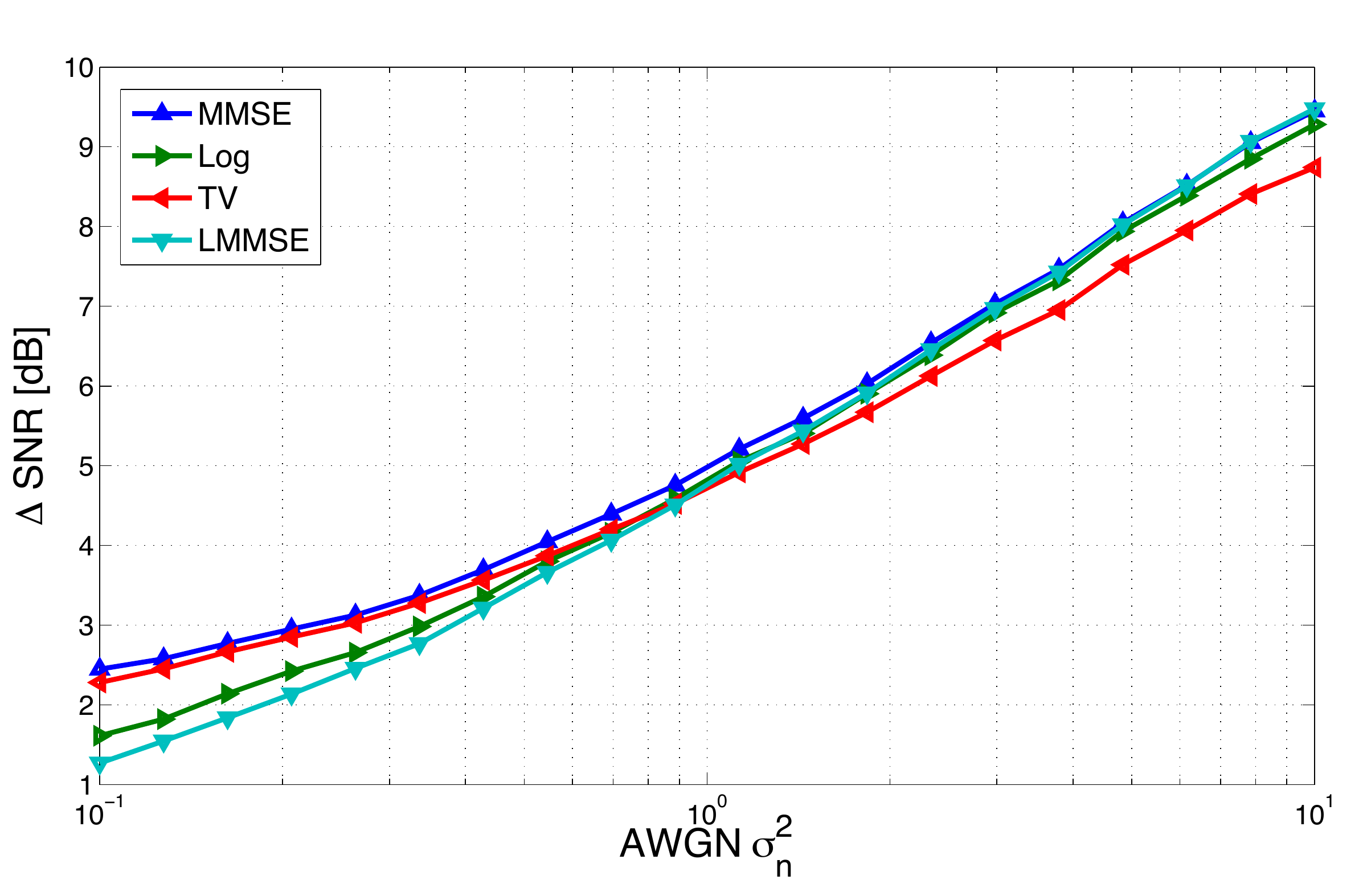}
			\caption{SNR improvement vs. variance of the additive noise for Gaussian compound Poisson innovations. The denoising methods are: MMSE estimator, Log regularization, TV regularization, and LMMSE estimator.}
			\label{fig:GaussianCP_SNRimprovement}
		\end{minipage}
	\end{figure}
}

We show in Figure \ref{fig:Gaussian_SNRimprovement} the SNR improvement of a Gaussian process after denoising by the four methods. Since the LMMSE and MMSE methods are equivalent in the Gaussian case, only the MMSE curve obtained from the message-passing algorithm is plotted. As expected, the MMSE method outperforms the TV and Log regularization techniques. The counter intuitive observation is that Log, which includes a nonconvex penalty function, performs better than TV. Another advantage of the Log regularizer is that it is differentiable and quadratic around the origin.

\ifthenelse{\equal{\Ncol}{two}}{
	\begin{figure}
	\centering
	\includegraphics[width=8.5cm]{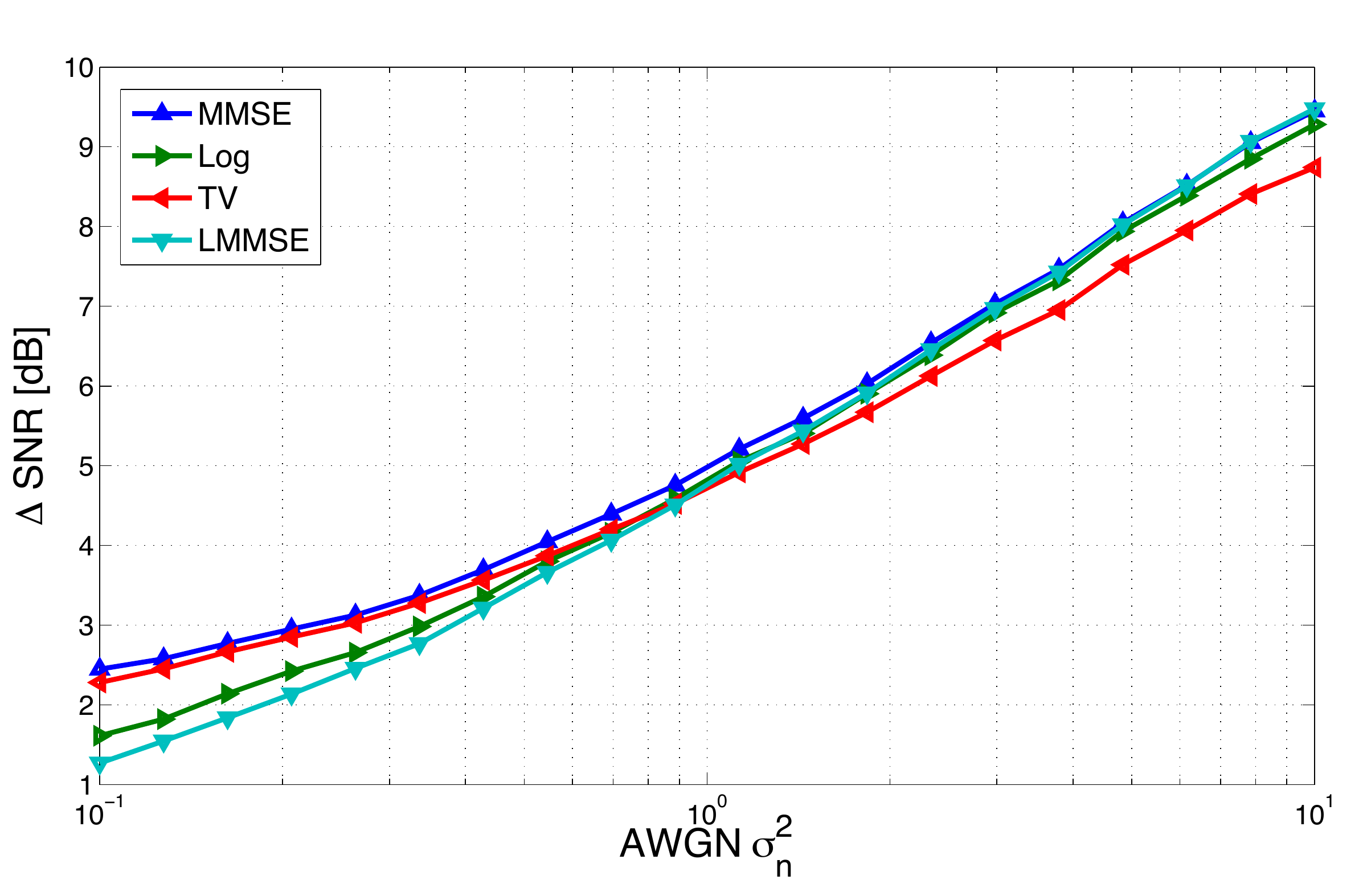}
	\caption{SNR improvement vs. variance of the additive noise for Gaussian compound Poisson innovations. The denoising methods are: MMSE estimator, Log regularization, TV regularization, and LMMSE estimator.}
	\label{fig:GaussianCP_SNRimprovement}
	\end{figure}
}{}

A similar scenario is repeated in Figure \ref{fig:GaussianCP_SNRimprovement} for the compound-Poisson innovation with $\lambda=0.6$ and Gaussian amplitudes (zero-mean and $\sigma=1$). As mentioned in Section \ref{subsec:LevyMAPdenoising}, since the pdf of the increments contains a mass probability at $x=0$, the MAP estimator selects the all-zero signal as the most probable choice. In Figure \ref{fig:GaussianCP_SNRimprovement}, this trivial estimator is excluded from the comparison. 
It can be observed that the performance of the MMSE denoiser, which is considered to be the gold standard, is very close to that of the TV regularization method at low noise powers where the source sparsity dictates the structure. This is consistent with what was predicted in \cite{Gribonval2011a}. Meanwhile, it performs almost as well as the LMMSE method at large noise powers. 
There, the additive Gaussian noise is the dominant term and the statistics of the noisy signal is mostly determined by the Gaussian constituent, which is matched to the LMMSE method. Excluding the MMSE method, none of the other three outperforms another one for the entire range of noise.

\ifthenelse{\equal{\Ncol}{two}}{
	\begin{figure}
	\centering
	\includegraphics[width=8.5cm]{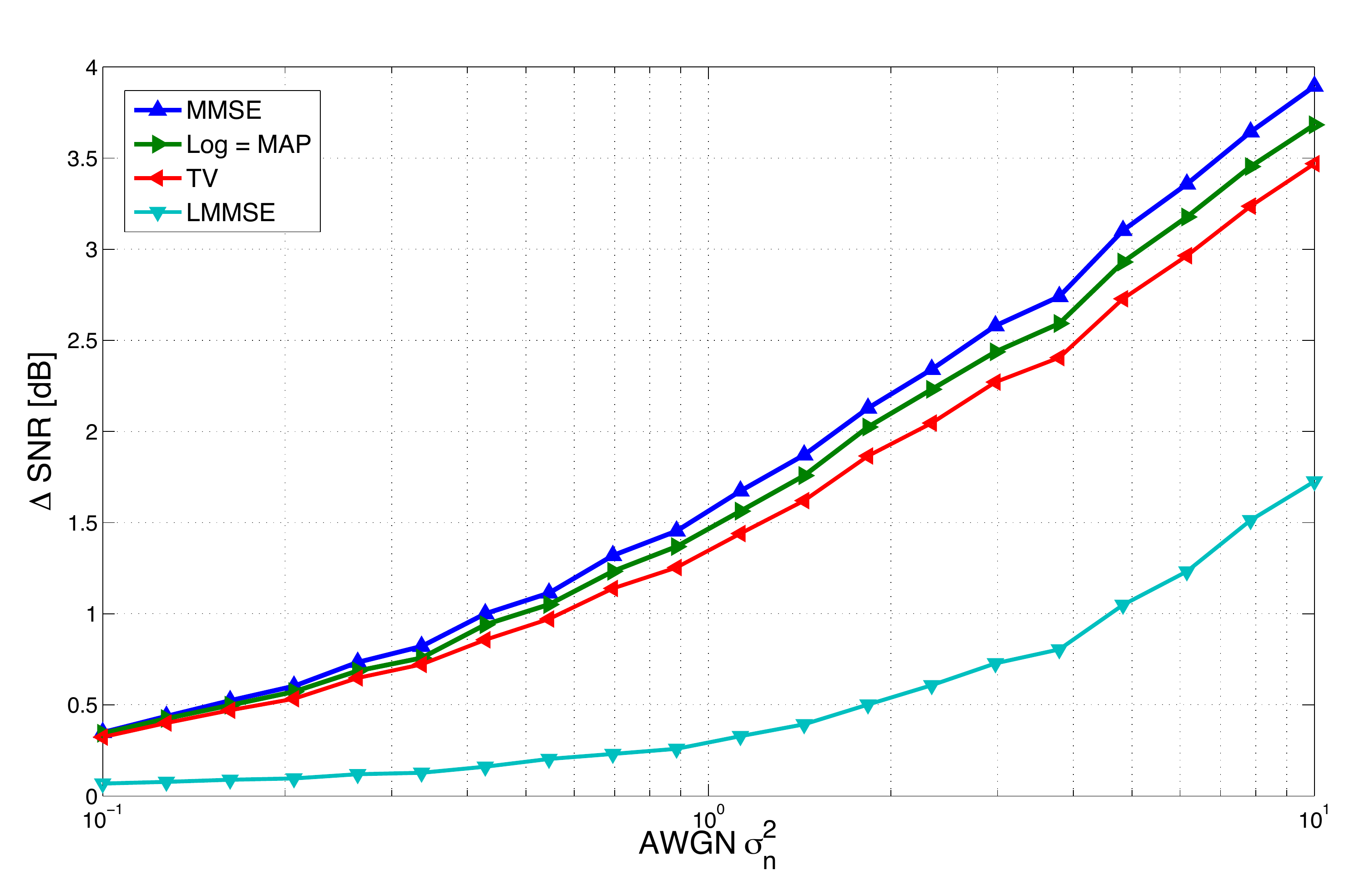}
	\caption{SNR improvement vs. variance of the additive noise for Cauchy ($\SO$-stable with $\SO=1$) innovations. The denoising methods are: MMSE estimator, Log regularization (which is equivalent to MAP here), TV regularization, and LMMSE estimator.}
	\label{fig:Cauchy_SNRimprovement}
	\end{figure}
}{
	\begin{figure}[t]
		\begin{minipage}[b]{0.5\linewidth}
			\centering
			\includegraphics[width=8.5cm]{Cauchy.pdf}
			\caption{SNR improvement vs. variance of the additive noise for Cauchy ($\SO$-stable with $\SO=1$) innovations. The denoising methods are: MMSE estimator, Log regularization (which is equivalent to MAP here), TV regularization, and LMMSE estimator.}
		\label{fig:Cauchy_SNRimprovement}
		\end{minipage}
		\hspace{0.5cm}
		\begin{minipage}[b]{0.5\linewidth}
			\centering
			\includegraphics[width=8.5cm]{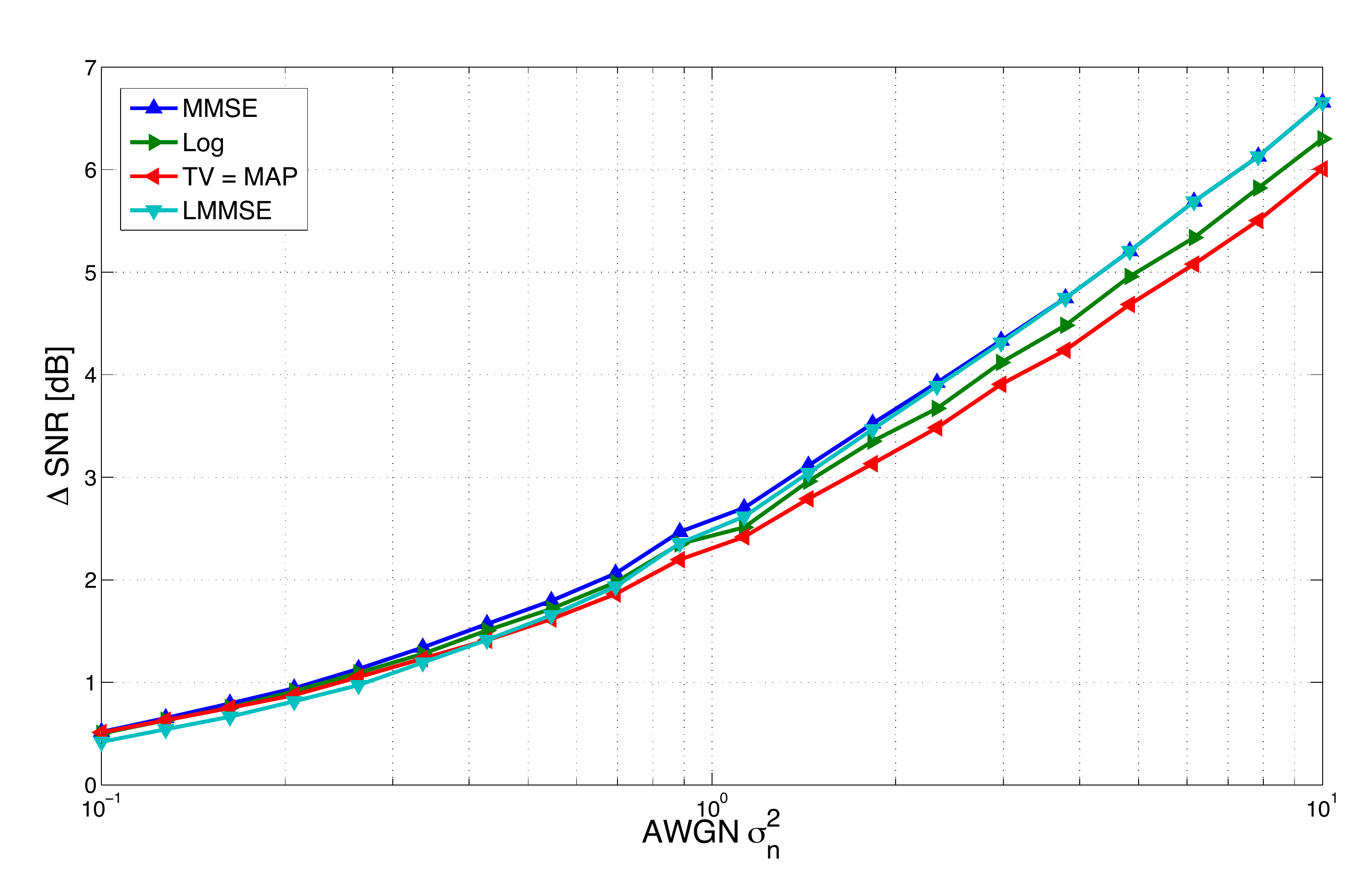}
			\caption{SNR improvement vs. variance of the additive noise for Laplace-type innovations. The denoising methods are: MMSE estimator, Log regularization, TV regularization (which is equivalent to MAP here),  and LMMSE estimator.}
	\label{fig:Laplace_SNRimprovement}
		\end{minipage}
	\end{figure}
}

Heavy-tail distributions such as $\alpha$-stables produce sparse or compressible sequences. With high probability, their realizations consist of a few large peaks and many insignificant samples. Since the convolution of a heavy-tail pdf with a Gaussian pdf is still heavy-tail, the noisy signal looks sparse even at large noise powers. The poor performance of the LMMSE method observed in Figure \ref{fig:Cauchy_SNRimprovement} for Cauchy distributions confirms this characteristic. The pdf of the Cauchy distribution, given by $\frac{1}{\pi(1+x^2)}$, is in fact the symmetric $\alpha$-stable distribution with $\alpha=1$. The Log regularizer corresponds to the MAP estimator of this distribution while there is no direct link between the TV regularizer and the MAP or MMSE criteria. The SNR improvement curves in Figure \ref{fig:Cauchy_SNRimprovement} indicate that the MMSE and Log (MAP) denoisers for this sparse process perform similarly (specially at small noise powers) and outperform the corresponding $\ell_1$-norm regularizer (TV).

\ifthenelse{\equal{\Ncol}{two}}{
	\begin{figure}
	\centering
	\includegraphics[width=8.5cm]{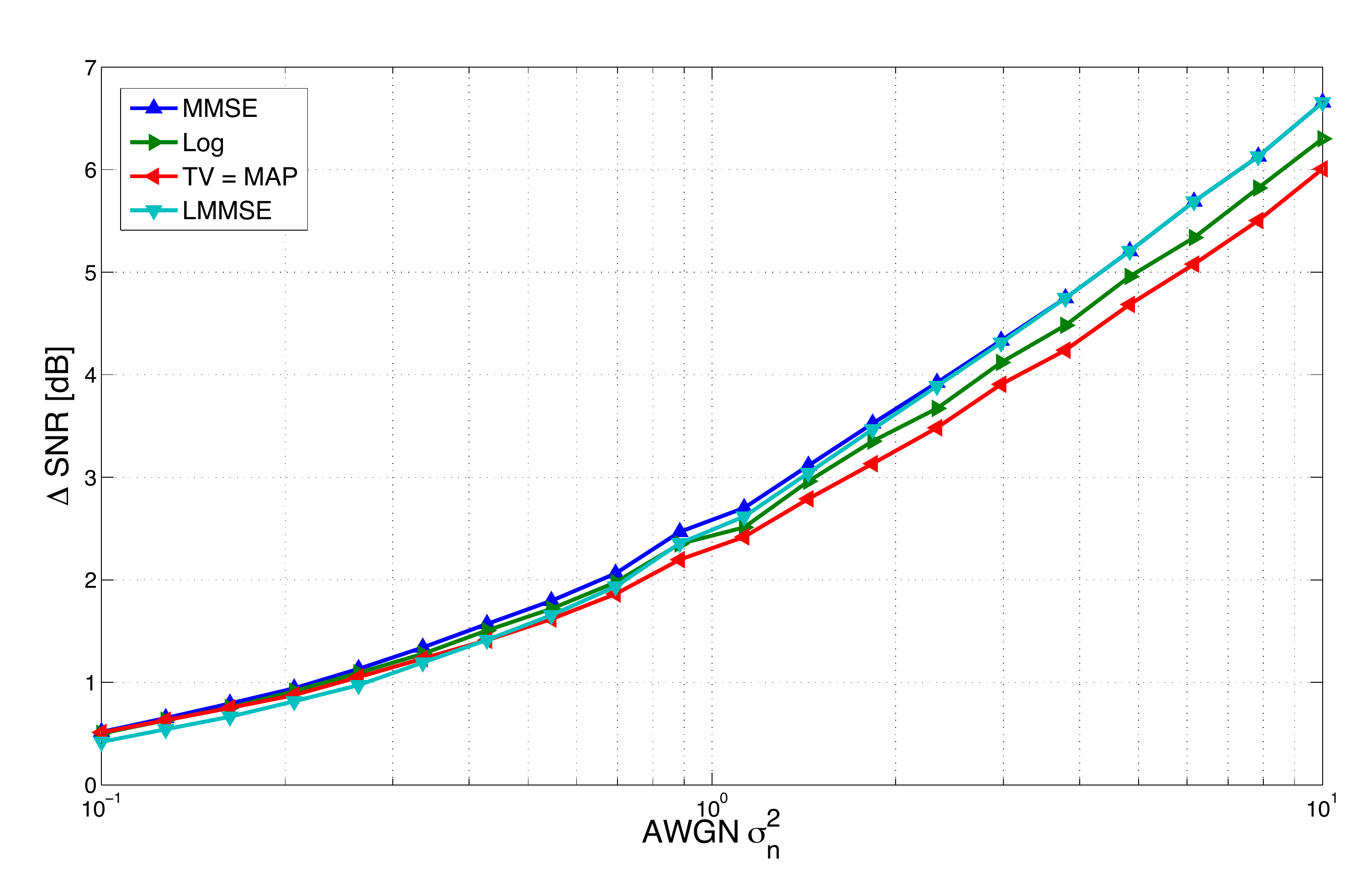}
	\caption{SNR improvement vs. variance of the additive noise for Laplace-type innovations. The denoising methods are: MMSE estimator, Log regularization, TV regularization (which is equivalent to MAP here),  and LMMSE estimator.}
	\label{fig:Laplace_SNRimprovement}
	\end{figure}
}{}

In the final scenario, we consider innovations with $\bbTwo=0$ and $v(a)=\frac{\e^{-|a|}}{|a|}$. This results in finite differences obtained at $T=1$ that follow a Laplace distribution  (see Appendix \ref{sec:AppTV_MAP}). 
Since the MAP denoiser for this process coincides with TV regularization, sometimes the Laplace distribution has been considered to be a sparse prior. However, it is proved in \cite{Amini2011}, \cite{Gribonval2012} that the realizations of a sequence with Laplace prior are not compressible, almost surely. The curves presented in Figure \ref{fig:Laplace_SNRimprovement} show that TV is a good approximation of the MMSE method only in light-noise conditions. For moderate to large noise, the LMMSE method is better than TV.


\section{Conclusion}\label{sec:Conc}

In this paper, we studied continuous-time stochastic processes where the process is defined by applying a linear operator on a white innovation process. For specific types of innovation, the procedure results in sparse processes. We derived a factorization of the joint posterior distribution for the noisy samples of the broad family ruled by fixed-coefficient stochastic differential equations. The factorization allows us to efficiently apply statistical estimation tools. 
A consequence of our pdf factorization is that it gives us access to the MMSE estimator. It can then be used as a gold standard for evaluating the performance of regularization techniques. This enables us to replace the MMSE method with a more-tractable and computationally efficient regularization technique matched to the problem without compromising the performance.
We then focused on L\'evy processes as a special case for which we studied  the denoising and interpolation problems using MAP and MMSE methods. We also compared these methods with the popular regularization techniques for the recovery of sparse signals, including the $\ell_1$ norm (e.g., TV regularizer) and the Log regularization approaches. Simulation results showed that we can almost achieve the MMSE performance by tuning the regularization technique to the type of innovation and the power of the noise.
We have also developed a graphical user interface in MATLAB which generates realizations of stochastic processes with various types of innovation and allows the user to apply either the MMSE or variational methods to denoise the samples\footnote{The GUI is available at \url{http://bigwww.epfl.ch/amini/MATLAB_codes/SSS_GUI.zip}}.


\appendices

\section{Characteristic Forms}\label{sec:AppCharForm}
In Gelfand's theory of generalized random processes,  the process is defined through its inner product with a space of test functions, rather than point values. For a random process $\RP$ and an arbitrary test function $\varphi$ chosen from a given space, the characteristic form is  defined as the characteristic function of the random variable $\ip=\langle \RP,\varphi\rangle$ and given by
\ifthenelse{\equal{\Ncol}{two}}{
	\begin{eqnarray}
	\CH_{\RP}(\varphi)&=&\EV\big\{\e^{-\j\langle \RP,\varphi\rangle}\big\}=\int_{\mathbb{R}}p_{\ip}(\ipl)\e^{-\j\ipl} \d\ipl\nonumber\\
	&=&\mathcal{F}_{\ipl}\big\{p_{\ip}(\ipl)\big\}(1).
	\end{eqnarray}
}{
	\begin{eqnarray}
	\CH_{\RP}(\varphi) = \EV\big\{\e^{-\j\langle \RP,\varphi\rangle}\big\}=\int_{\mathbb{R}}p_{\ip}(\ipl)\e^{-\j\ipl} \d\ipl = \mathcal{F}_{\ipl}\big\{p_{\ip}(\ipl)\big\}(1).
	\end{eqnarray}
}
As an example, let $w_{_{\rm G}}$ be a normalized white Gaussian noise and let $\varphi$ be an arbitrary function in $L_2(\mathbb{R})$. It is well-known that $\langle w_{_{\rm G}},\varphi\rangle$ is a zero-mean Gaussian random variable with variance $\|\varphi\|_{L_2}^2$. Thus, in this example we have that
\begin{eqnarray}
\CH_{w_{_{\rm G}}}(\varphi)=\e^{-\frac{1}{2}\|\varphi\|_{L_2}^2}.
\end{eqnarray}

An interesting property of the characteristic forms is that they help determine the joint probability density functions for arbitrary finite dimensions as
\ifthenelse{\equal{\Ncol}{two}}{
	\begin{eqnarray}\label{eq:ProbFourierCharForm}
	\CH_{\RP}\big(\sum_{i=1}^k\omega_i\varphi_i\big) &=&\EV\big\{\e^{-\j\langle \RP,	\sum_{i=1}^k\omega_i\varphi_i\rangle}\big\}\nonumber\\
	&=&\EV\big\{\e^{-\j\sum_{i=1}^k\omega_i \ip_i}\big\} \nonumber\\
	&=&\mathcal{F}_{\mathbf{\ipl}}\big\{p_{\ip}(\mathbf{\ipl})\big\}(\omega_1,\dots,\omega_k),
	\end{eqnarray}
}{
	\begin{eqnarray}\label{eq:ProbFourierCharForm}
	\CH_{\RP}\big(\sum_{i=1}^k\omega_i\varphi_i\big) &=&\EV\big\{\e^{-\j\langle \RP,	\sum_{i=1}^k\omega_i\varphi_i\rangle}\big\} = \EV\big\{\e^{-\j\sum_{i=1}^k\omega_i \ip_i}\big\} \nonumber\\
	&=&\mathcal{F}_{\mathbf{\ipl}}\big\{p_{\ip}(\mathbf{\ipl})\big\}(\omega_1,\dots,\omega_k),
	\end{eqnarray}
}
where $\{\varphi_i\}_i$ are test functions and $\{\omega_i\}$ are scalars. Equation (\ref{eq:ProbFourierCharForm}) shows that an inverse Fourier transform of the characteristic form can yield the desired pdf. Beside  joint distributions, characteristic forms are useful for generating moments too:
\ifthenelse{\equal{\Ncol}{two}}{
	\begin{small}
	\begin{eqnarray}
	&&\frac{\partial^{n_1+\dots+n_k}}{\partial\omega_1^{n_1}\cdots\partial\omega_k^{n_k}}\CH_{\RP}\big(\sum_{i=1}^k\omega_i \varphi_i\big)\Big|_{\omega_1=\dots=\omega_k=0}\nonumber\\
	&&=(-\j)^{n_1+\dots+n_k}\int_{\mathbb{R}^k}\ipl_1^{n_1}\cdots \ipl_k^{n_k} p_{\ip}(\ipl_1,\dots,\ipl_k)\d \ipl_1\cdots \d \ipl_k\nonumber\\
	&&=(-\j)^{n_1+\dots+n_k}\EV\big\{\ipl_1^{n_1}\cdots \ipl_k^{n_k}\big\}.
	\end{eqnarray}
	\end{small}
}{
	\begin{eqnarray}
	\frac{\partial^{n_1+\dots+n_k}}{\partial\omega_1^{n_1}\cdots\partial\omega_k^{n_k}}\CH_{\RP}\big(\sum_{i=1}^k\omega_i \varphi_i\big)\Big|_{\omega_1=\dots=\omega_k=0}&=&(-\j)^{n_1+\dots+n_k}\int_{\mathbb{R}^k}y_1^{n_1}\cdots y_k^{n_k}p_y(y_1,\dots,y_k)\d y_1\cdots \d y_k\nonumber\\
	&=&(-\j)^{n_1+\dots+n_k}\EV\big\{y_1^{n_1}\cdots y_k^{n_k}\big\}.
	\end{eqnarray}
}

Note that the definition of random processes through characteristic forms includes the classical definition based on the point values by choosing Diracs as the test functions (if possible).

Except for the stable processes, it is usually hard to find the distributions of linear transformations of a process. However, there exists a simple relation between the characteristic forms: Let $\RP$ be a random process and define $\xi=\L \RP$, where $\L$ is a linear operator. Also denote the adjoint of $\L$ by $\L^{*}$. Then, one can write
\ifthenelse{\equal{\Ncol}{two}}{
	\begin{eqnarray}
	\CH_{\xi}(\varphi)&=&\EV\big\{\e^{-\j\langle\L \RP,\varphi\rangle}\big\} =\EV\big\{\e^{-\j\langle \RP,\L^{*}\varphi\rangle}\big\} \nonumber\\
	&=&\CH_{\RP}(\L^{*} \varphi).
	\end{eqnarray}
}{
	\begin{eqnarray}
	\CH_{\xi}(\varphi) = \EV\big\{\e^{-\j\langle\L \RP,\varphi\rangle}\big\} =\EV\big\{\e^{-\j\langle \RP,\L^{*}\varphi\rangle}\big\} = \CH_{\RP}(\L^{*} \varphi).
	\end{eqnarray}
}
Now it is easy to extract the probability distribution of $\xi$ from its characteristic form.


\section{Specification of $n$th-order Shaping Kernels}\label{sec:AppShapingKernel}

To show the existence of a  kernel $h$ for the $n$th-order differential operator $\L=\ARC_n\prod_{i=1}^n (\D-r_i\I)$, we define
\ifthenelse{\equal{\Ncol}{two}}{
	\begin{eqnarray}\label{eq:hiDef}
	h_i(x,\tau) = \left\{ \begin{array}{ll}
	\e^{r_i(x-\tau)} \big(\bfOne_{[0,\infty[}(x-\tau) & \\
	\phantom{\e^{r_i(x-\tau)} \big(} - \bfOne_{[0,\infty[}(\bar{x}_i-\tau)\big), & \Re r_i=0, \\
	\e^{r_i(x-\tau)} \bfOne_{[0,\infty[}(x-\tau), & \Re r_i<0,\\
	\end{array}\right.
	\end{eqnarray}
}{
	\begin{eqnarray}\label{eq:hiDef}
	h_i(x,\tau) = \left\{ \begin{array}{ll}
	\e^{r_i(x-\tau)} \big(\bfOne_{[0,\infty[}(x-\tau) - \bfOne_{[0,\infty[}(\bar{x}_i-\tau)\big), & \Re r_i=0, \\
	\e^{r_i(x-\tau)} \bfOne_{[0,\infty[}(x-\tau), & \Re r_i<0,\\
	\end{array}\right.
	\end{eqnarray}
}
where $\bar{x}_i$ are nonpositive fixed real numbers. It is not hard to check that $h_i$ satisfies the conditions (i)-(iii) for the operator $\L_i=\D-r_i\I$. Next, we combine $h_i$ to form a proper kernel for $\LI$ as
\begin{eqnarray}
h(x,\tau)= \frac{1}{\ARC_n} \int_{\R^{n-1}} \prod_{i=1}^{n}h_i(\tau_{i+1},\tau_i) \prod_{i=2}^n \d\tau_i \Big|_{\substack{\tau_1=\tau ~~\;\\ \tau_{n+1}=x}} .
\end{eqnarray}
By relying on the fact that the $h_i$ satisfy conditions (i)-(iii), it is possible to prove by induction that $h$ also satisfies (i)-(iii). Here, we only provide the main idea for proving (i). We use the factorization $\L=\ARC_n \L_1\cdots\L_n$ and sequentially apply every $\L_i$ on $h$. The starting point $i=n$ yields
\ifthenelse{\equal{\Ncol}{two}}{
	\begin{eqnarray}
	\L _n h(x,\tau) &=& \int_{\R^{n-1}} \delta(x-\tau_n)\frac{\prod_{i=1}^{n}h_i(\tau_{i+1},\tau_i)  \d\tau_i }{\ARC_n h_n(\tau_{n+1},\tau_n)}\Big|_{\substack{\tau_1=\tau}} \nonumber\\
	&=&\frac{1}{\ARC_n} \int_{\R^{n-2}} \prod_{i=1}^{n-1}h_i(\tau_{i+1},\tau_i) \prod_{i=2}^{n-1} \d\tau_i \Big|_{\substack{\tau_1=\tau\,\\ \tau_{n}=x}}.
	\end{eqnarray}
}{
	\begin{eqnarray}
	\L _n h(x,\tau) = \int_{\R^{n-1}} \delta(x-\tau_n)\frac{\prod_{i=1}^{n}h_i(\tau_{i+1},\tau_i)  \d\tau_i }{\ARC_n h_n(\tau_{n+1},\tau_n)}\Big|_{\substack{\tau_1=\tau}} =\frac{1}{\ARC_n} \int_{\R^{n-2}} \prod_{i=1}^{n-1}h_i(\tau_{i+1},\tau_i) \prod_{i=2}^{n-1} \d\tau_i \Big|_{\substack{\tau_1=\tau\,\\ \tau_{n}=x}}.
	\end{eqnarray}
}
Thus, $\L_n h(x,\tau)$ has the same form as $h$ with $n$ replaced by $n-1$. By continuing the same procedure, we finally arrive at $\L_1 h_1(x,t)$, which is equal to $\delta(x-\tau)$.


\section{Proof of Theorem \ref{theo:s2dprob}}\label{sec:AppTheoFactorization}

For the sake of simplicity in the notations, for $\theta\geq n$ we define
\ifthenelse{\equal{\Ncol}{two}}{
	\begin{small}
	\begin{eqnarray}
	\mathbf{u}[\theta]=\left[\begin{array}{c}
	u_T[\theta]\\
	u_T[\theta-1]\\
	\vdots\\
	u_T[n]\\
	s_T[n-1]\\
	s_T[n-2]\\
	\vdots\\
	s_T[0]
	\end{array}\right].
	\end{eqnarray}
	\end{small}
}{
	\begin{eqnarray}
	\mathbf{u}[\theta]=\big[ u_T[\theta] \; u_T[\theta-1] \; \dots \;	u_T[n] \; s_T[n-1] \; 	s_T[n-2] \; \dots \; s_T[0]\big]^T.
	\end{eqnarray}
}

Since the $u_T[i]$ are linear combinations of $s_T[i]$, the $\big((\AI+1)\times 1\big)$ vector $\mathbf{u}[\AI]$ can be linearly expressed in terms of $s_T[i]$ as
\begin{eqnarray}\label{eq:MatrixEq}
\mathbf{u}[\AI]=\mathbf{D}_{(\AI+1)\times (\AI+1)}\left[\begin{array}{c}
s_T[\AI]\\
s_T[\AI-1]\\
\vdots \\
s_T[0]
\end{array}\right],
\end{eqnarray}
where $\mathbf{D}_{(\AI+1)\times (\AI+1)}$ is an upper-triangular matrix defined by 
\begin{align}
\left[
\begin{array}{c}
\begin{array}{ccccccc}
d_T[0] 	& d_T[1] 	& \cdots 	& d_T[n] 		& 0			& \cdots 	& 0 			\\
0 			& d_T[0] 	& \cdots 	& d_T[n-1] 	& d_T[n] 	& \cdots 	& 0			\\
\vdots	&   			&  			& 	 				& 				& 				& \vdots	\\
0 			& 0 			& \cdots 	& 0 				& d_T[0]	& \cdots 	& d_T[n] 	\\
\end{array} \\
--------------------- \\
\begin{array}{cccc}
\phantom{--} & \mathbf{0}_{n\times (\AI+1-n)} 	&  \phantom{-----}	 &  \mathbf{I}_{n\times n}  			\\
\end{array}
\end{array}\right].
\end{align}

Since $d_T[0]\neq 0$, none of the diagonal elements of the upper-triangular matrix $\mathbf{D}_{(\AI+1)\times (\AI+1)}$ is zero. Thus, the matrix is invertible because $\det \mathbf{D}_{(\AI+1)\times (\AI+1)} =(d_T[0])^{\AI+1-n}$. Therefore, we have that
\begin{eqnarray}\label{eq:JointS1}
p_{s,u}\big(\mathbf{u}[\AI]\big)=\frac{p_s\big(s_T[\AI],\dots,s_T[0]\big)}{\big|d_T[0]\big|^{\AI+1-n}}.
\end{eqnarray}

A direct consequence of Lemma \ref{lemma:indep} is that, for $\AI\geq 2n-1$, we obtain
\begin{eqnarray}
p_{s,u}\big(u_T[\AI] \, \big| \, \mathbf{u}[\AI-1]\big)=p_{u}\Big(u_T[\AI] \, \Big| \, \big\{u_T[\AI-i]\big\}_{i=1}^{n-1} \Big)
\end{eqnarray}
which, in conjunction with Bayes' rule, yields
\ifthenelse{\equal{\Ncol}{two}}{
	\begin{eqnarray}\label{eq:OneTermBays}
	\frac{p_{s,u}\big(\mathbf{u}[\AI]\big)}{p_{s,u}\big(\mathbf{u}[\AI-1]\big)}&=&p_{s,u}\big(u_T[\AI] \, \big| \, \mathbf{u}[\AI-1]\big)\nonumber\\
	&=&p_{u}\Big(u_T[\AI] \, \Big| \, \big\{u_T[\AI-i]\big\}_{i=1}^{n-1} \Big).
	\end{eqnarray}
}{
	\begin{eqnarray}\label{eq:OneTermBays}
	\frac{p_{s,u}\big(\mathbf{u}[\AI]\big)}{p_{s,u}\big(\mathbf{u}[\AI-1]\big)}&=&p_{s,u}\big(u_T[\AI] \, \big| \, \mathbf{u}[\AI-1]\big) = p_{u}\Big(u_T[\AI] \, \Big| \, \big\{u_T[\AI-i]\big\}_{i=1}^{n-1} \Big).
	\end{eqnarray}
}

By multiplying equations of the form (\ref{eq:OneTermBays}) for $\AI=2n-1,\dots,k$, we get
\begin{small}
\begin{eqnarray}\label{eq:AllBays}
\frac{p_{s,u}\big(\mathbf{u}[k]\big)}{p_{s,u}\big(\mathbf{u}[2n-2]\big)}&=&\prod_{\AI=2n-1}^{k}p_{u}\Big(u_T[\AI] \, \Big| \, \big\{u_T[\AI-i]\big\}_{i=1}^{n-1}\big).
\end{eqnarray}
\end{small}

It is now easy to complete the proof by substituting the numerator and denominator of the left-hand side in (\ref{eq:AllBays}) by the equivalent forms suggested by (\ref{eq:JointS1}).


\section{Proof of Theorem \ref{theo:Cond_d_CharForm}}\label{sec:AppTheoPDF}

As developed in Appendix \ref{sec:AppCharForm}, the characteristic form can be used to generate the joint probability density functions. To use (\ref{eq:ProbFourierCharForm}), we need to represent $u_T[i]$ as inner-products with the white process. This is already available from (\ref{eq:uTWN}). This yields

\ifthenelse{\equal{\Ncol}{two}}{
	\begin{small}
	\begin{eqnarray}\label{eq:ProbCharGen3}
	&p_u\big(\{u_T[\AI-i]\}_{i=0}^{k}\big)=&\nonumber\\
	&\mathcal{F}^{-1}_{\{\omega_{i}\}}\Big\{\CH_{w}\big(\sum_{i=0}^{k}\omega_i\SP(\AI T-iT-\cdot)\big)\Big\}\big(\{u_T[\AI-i]\}_{i=0}^{k}\big).&
	\end{eqnarray}
	\end{small}
}{
	\begin{eqnarray}\label{eq:ProbCharGen3}
	p_u\big(\{u_T[\AI-i]\}_{i=0}^{k}\big) = \mathcal{F}^{-1}_{\{\omega_{i}\}}\Big\{\CH_{w}\big(\sum_{i=0}^{k}\omega_i\SP(\AI T-iT-\cdot)\big)\Big\}\big(\{u_T[\AI-i]\}_{i=0}^{k}\big).
	\end{eqnarray}
}

From (\ref{eq:GelfandGeneric}), we have
\ifthenelse{\equal{\Ncol}{two}}{
	\begin{eqnarray}\label{eq:CPPd1}
	&& \CH_{w}\big(\sum_{i=0}^{k}\omega_i\SP(\AI T-iT-\cdot)\big) \nonumber\\
	&&~~~~~~~~~~ = \e^{\int_{\mathbb{R}} f_w\big(\sum_{i=0}^{k}\omega_i\SP(\AI T-iT-x)\big)\d x  }\nonumber\\
	&&~~~~~~~~~~ = \e^{\int_{\mathbb{R}} f_w\big(\sum_{i=0}^{k}\omega_i\SP(x-iT)\big)\d x  }.
	\end{eqnarray}
}{
	\begin{eqnarray}\label{eq:CPPd1}
	\CH_{w}\big(\sum_{i=0}^{k}\omega_i\SP(\AI T-iT-\cdot)\big) = \e^{\int_{\mathbb{R}} f_w\big(\sum_{i=0}^{k}\omega_i\SP(\AI T-iT-x)\big)\d x  }= \e^{\int_{\mathbb{R}} f_w\big(\sum_{i=0}^{k}\omega_i\SP(x-iT)\big)\d x  }.
	\end{eqnarray}
}

Using (\ref{eq:Ifunc}), it is now easy to verify that
\ifthenelse{\equal{\Ncol}{two}}{
	\begin{small}
	\begin{eqnarray}
	&\CH_{w}\big(\sum_{i=0}^{n-1}\omega_i\SP(\AI T-iT-\cdot)\big) =\e^{I_{w,\SP}(\omega_0,\dots,\omega_{n-1}) },& \nonumber\\
	&\CH_{w}\big(\sum_{i=1}^{n-1}\omega_i\SP(\AI T-iT-\cdot)\big) = \e^{I_{w,\SP}(0,\omega_1,	\dots,\omega_{n-1}) } . &
	\end{eqnarray}
	\end{small}
}{
	\begin{eqnarray}
	\begin{array}{ccc}
	\CH_{w}\big(\sum_{i=0}^{n-1}\omega_i\SP(\AI T-iT-\cdot)\big) & = & \e^{I_{w,\SP}(\omega_0,\dots,\omega_{n-1}) }, \\
	\CH_{w}\big(\sum_{i=1}^{n-1}\omega_i\SP(\AI T-iT-\cdot)\big) & = & \e^{I_{w,\SP}(0,\omega_1,	\dots,\omega_{n-1}) }. \\
	\end{array}
	\end{eqnarray}
}
The only part left to mention before completing the proof is that
\begin{eqnarray}
p_{u}\Big(u_T[\AI]~\Big|~\big\{u_T[\AI-i]\big\}_{i=1}^{n-1}\Big)= \frac{ p_{u}\Big(\big\{u_T[\AI-i]\big\}_{i=0}^{n-1} \Big)}{ p_{u}\Big(\big\{u_T[\AI-i]\big\}_{i=1}^{n-1} \Big) }. 
\end{eqnarray}


\section{When does TV regularization meet MAP?}\label{sec:AppTV_MAP}
The TV-regularization technique is one of the successful methods in denoising. 
Since the TV penalty is separable with respect to first-order finite differences, its interpretation as a MAP estimator is valid only for a L\'evy process. Moreover, the MAP estimator of a L\'evy process coincides with TV regularization only if $\Psi_T(x)=-\log p_u(x)=\gamma |x|+\eta$, where $\gamma$ and $\eta$ are constants such that $\gamma>0$. This condition implies that $p_u$ is the  Laplace pdf  $p_u(x)=\frac{\gamma}{2}\e^{-\gamma |x|}$. This pdf is a valid distribution for the first-order finite differences of the L\'evy process characterized by the innovation with $\bbOne=\bbTwo=0$ and $v(a)=\frac{\e^{-\gamma |a|}}{|a|}$ because
\ifthenelse{\equal{\Ncol}{two}}{
	\begin{eqnarray}
	f_l(\omega) &=& \int_{\mathbb{R}}\big(\e^{\j\omega a}-1\big)\frac{\e^{-\gamma |a|}}{|a|}\d a \nonumber\\
	&=& 2\int_{0}^{\infty}\big(\cos(\omega a)-1\big)\frac{\e^{-\gamma a}}{a}\d a.
	\end{eqnarray}
}{
	\begin{eqnarray}
	f_l(\omega) = \int_{\mathbb{R}}\big(\e^{\j\omega a}-1\big)\frac{\e^{-\gamma |a|}}{|a|}\d a =  2\int_{0}^{\infty}\big(\cos(\omega a)-1\big)\frac{\e^{-\gamma a}}{a}\d a.
	\end{eqnarray}
}
Thus, we can write
\ifthenelse{\equal{\Ncol}{two}}{
	\begin{eqnarray}\label{eq:Diff_f_lap}
	\frac{\d}{\d\omega} f_l(\omega) &=& -2\int_{0}^{\infty}\sin(\omega a)\e^{-\gamma a}\d a\nonumber\\
	&=& \frac{-2\omega}{\gamma^2+\omega^2}.
	\end{eqnarray}
}{
	\begin{eqnarray}\label{eq:Diff_f_lap}
	\frac{\d}{\d\omega} f_l(\omega) = -2\int_{0}^{\infty}\sin(\omega a)\e^{-\gamma a}\d a = \frac{-2\omega}{\gamma^2+\omega^2}.
	\end{eqnarray}
}

\begin{figure}
\centering
\includegraphics[width=8.5cm]{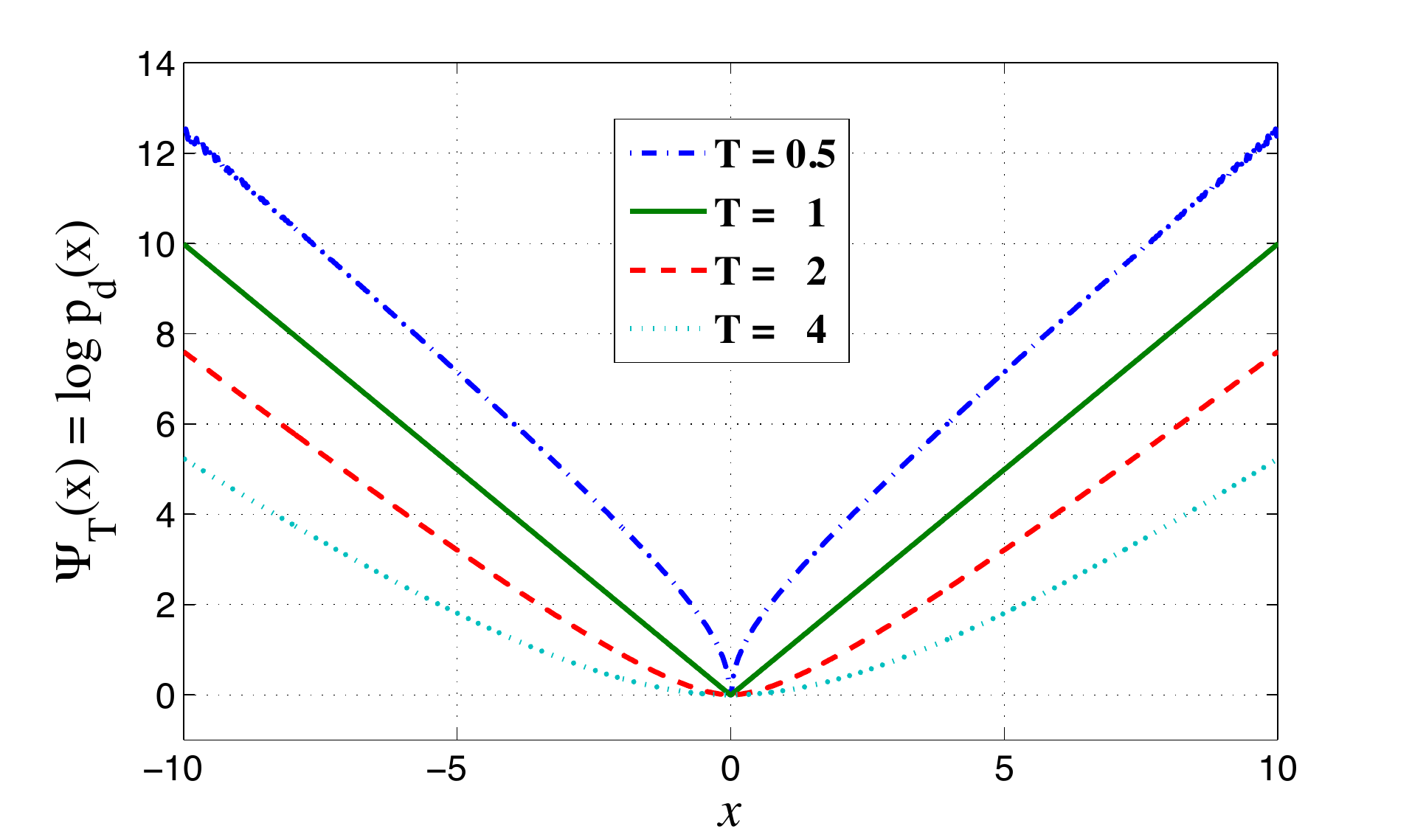}
\caption{The function $\Psi_T=-\log p_u$  for different values of $T$ after enforcing the curves to pass through the origin by applying a shift. For $T=1$, the density function $p_u$ follows a Laplace law. Therefore, the corresponding $\Psi_T$ is the absolute-value function.}
\label{fig:Psi_Laplace}
\end{figure}

By integrating (\ref{eq:Diff_f_lap}), we obtain that $f_l(\omega)=-\log (\gamma^2+\omega^2)+\eta$, where $\eta$ is a constant. The key point in finding this constant is the fact that $f_l(0)=0$, which results in $f_l(\omega)=\log \frac{\gamma^2}{\gamma^2+\omega^2}$. Now, for the sampling period $T$, Equation (\ref{eq:Levy_pdf}) suggests that
\ifthenelse{\equal{\Ncol}{two}}{
	\begin{eqnarray}
	p_u(x)&=&\mathcal{F}^{-1}_{\omega}\big\{\e^{Tf_l(\omega)}\big\}(x)=\mathcal{F}^{-1}_{\omega}\left\{\left(\frac{\gamma^2}	{\gamma^2+\omega^2}\right)^T\right\}(x) \nonumber\\
	&=& \frac{\gamma \, |\gamma x|^{T-\frac{1}{2}} K_{T-\frac{1}{2}}(|\gamma x|)}{\sqrt{\pi} \, 2^{T-\frac{1}{2}} \, \Gamma(T)},
	\end{eqnarray}
}{
	\begin{eqnarray}
	p_u(x)=\mathcal{F}^{-1}_{\omega}\big\{\e^{Tf_l(\omega)}\big\}(x)=\mathcal{F}^{-1}_{\omega}\left\{\left(\frac{\gamma^2}	{\gamma^2+\omega^2}\right)^T\right\}(x) = \frac{\gamma \, |\gamma x|^{T-\frac{1}{2}} K_{T-\frac{1}{2}}(|\gamma x|)}{\sqrt{\pi} \, 2^{T-\frac{1}{2}} \, \Gamma(T)},
	\end{eqnarray}
}
where $K_{t}(\cdot)$ is the modified Bessel function of the second kind. The latter probability density function is known as \emph{symmetric variance-gamma} or \emph{symm-gamma}. It is not hard to check that we obtain the desired Laplace distribution for $T=1$. However, this value of $T$ is the only one for which we observe this property. Should the sampling grid become finer or coarser, the MAP estimator would no longer coincide with TV regularization. We show in Figure \ref{fig:Psi_Laplace} the shifted $\Psi_T$ functions for various $T$ values for the aforementioned innovation where $\gamma=1$.

\bibliographystyle{IEEEtran} 
\bibliography{FRIdenoisingBib}

\end{document}